\documentclass{article}

\usepackage[final]{corl_2026} 
\usepackage{amsmath}
\usepackage{amssymb}
\usepackage{graphicx}
\usepackage{booktabs}
\usepackage{comment}
\usepackage[table]{xcolor}
\usepackage{float}
\usepackage{multirow}
\usepackage{tabularx}
\usepackage{array}
\usepackage{wrapfig}
\newcolumntype{Y}{>{\centering\arraybackslash}X}

\title{MoDex: Towards Universal Multi-Object Dexterous Grasping}

\author{
    \textbf{Jing Xiong}\textsuperscript{\normalfont 1,*} \quad 
    \textbf{Jiale Xie}\textsuperscript{\normalfont 1,*} \quad 
    \textbf{Haoqi Yuan}\textsuperscript{\normalfont 1,2} \quad 
    \textbf{Chaoyi Xu}\textsuperscript{\normalfont 1,2} \quad 
    \textbf{Zongqing Lu}\textsuperscript{\normalfont 1,2,\ensuremath{\dagger}} \\ 
    \textsuperscript{1}Peking University \quad 
    \textsuperscript{2}BeingBeyond \\ 
    \textsuperscript{*}Equal contribution \quad 
    \textsuperscript{\ensuremath{\dagger}}Corresponding author
}

\begin{document}
\maketitle

\begin{figure}[H]
    \centering
    \includegraphics[width=\linewidth]{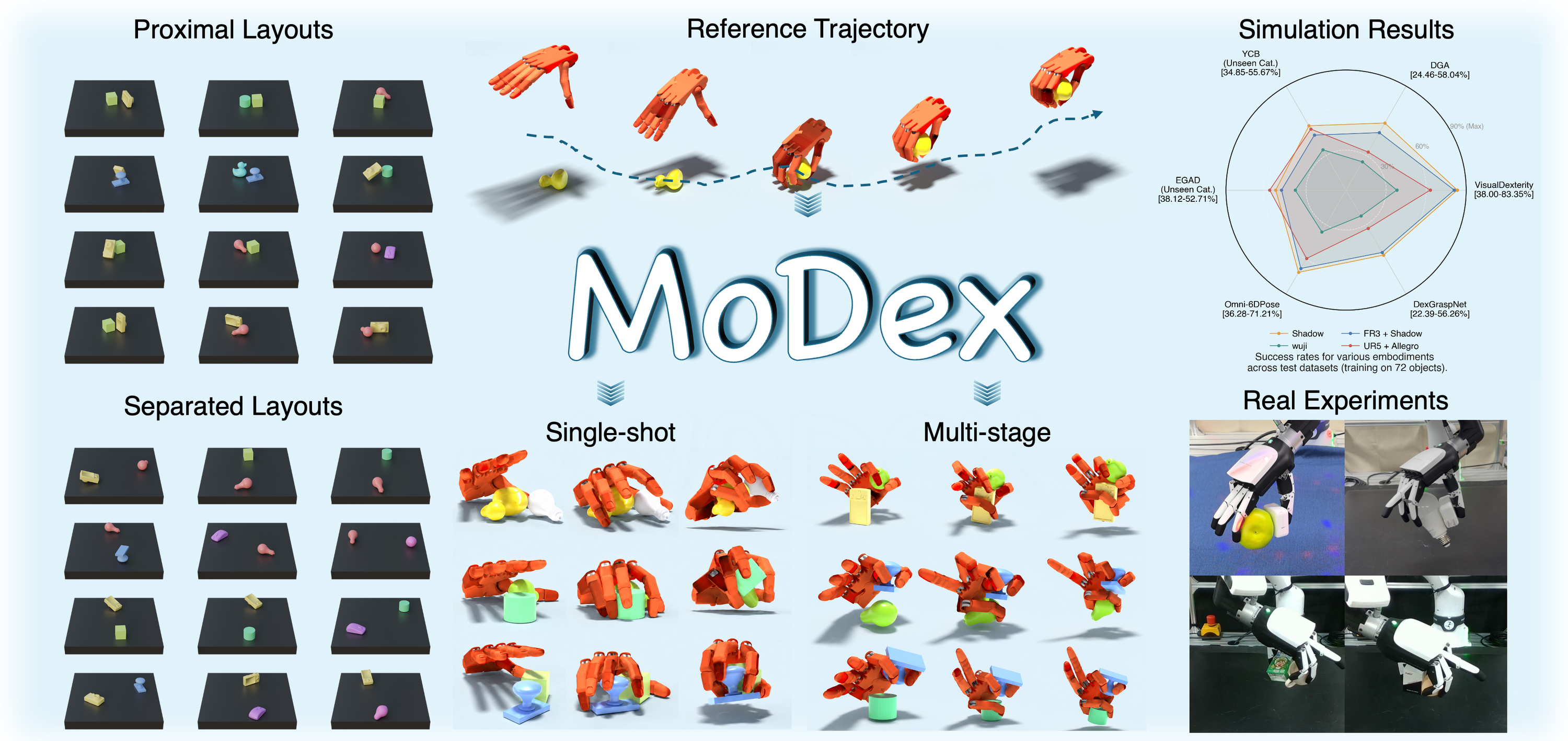}
    \caption{MoDex learns a multi-object dexterous grasping policy that generalizes to unseen object categories and layouts. It achieves state-of-the-art performance in simulation and zero-shot sim-to-real transfer on unseen real-world objects.}
    \label{fig:teaser}
\end{figure}

\begin{abstract}
Multi-object grasping is a fundamental yet challenging problem, constrained by data collection costs, varying spatial layouts, and the need for broad generalization. To address these challenges, we propose MoDex, a data-efficient framework that provides the first unified solution for universal multi-object dexterous grasping across diverse spatial layouts. MoDex accomplishes this by formulating the task as a single-step decision problem of learning residual corrections to a reference trajectory, and introducing a spatially adaptive dual-strategy that performs single-shot grasping for proximal objects and sequential interaction for separated objects. Extensive evaluations show that MoDex significantly outperforms baselines in proximal configurations, achieving 61.24\% success on unseen objects, and achieves high success rates in separated layouts. Further experiments demonstrate its generalization across robot embodiments, scalability and zero-shot sim-to-real deployment through a vision-based policy trained solely on simulation data.
Code and pretrained models are available at \href{https://github.com/Aurora-Xiong/MoDex.git}{\nolinkurl{https://github.com/Aurora-Xiong/MoDex}}.
\end{abstract}


\section{Introduction}
Multi-object grasping is a fundamental yet challenging problem in dexterous manipulation~\citep{agboh2022multi,harada1998kinematics,sun2022multi,yamada2015static,yao2023exploiting}. The intrinsic complexity of this task arises from three challenges: \textbf{data efficiency}, where diverse object combinations and layouts make large-scale data collection or grasp generation costly; \textbf{spatial adaptivity}, where proximal layouts may allow single-shot grasping while separated layouts require sequential interaction; \textbf{out-of-distribution robustness}, where the policy must transfer across object geometries, placements, embodiments, and real-world visual observations. An effective framework must therefore make sample-efficient use of limited supervision, adapt to varying spatial layouts, and generalize broadly.

Existing multi-object approaches typically rely on explicit grasp synthesis. MultiGrasp~\citep{li2024grasp} uses optimization or diffusion models to generate whole-hand grasps, while SeqGrasp~\citep{lu2025grasping} synthesizes stable sequential grasps using partial hand degrees of freedom. SeqMultiGrasp~\citep{he2025sequential} also studies sequential multi-object grasp synthesis. These generation-based approaches rely on explicit grasp synthesis, introducing additional computation as object combinations and layouts scale. Meanwhile, recent single-object dexterous grasping methods, such as UniDexGrasp++~\citep{wan2023unidexgrasp++}, ResDex~\citep{huang2025efficient}, and DemoGrasp~\citep{yuan2025demograsp}, have made progress in generalizing across diverse objects, but they are not directly applicable to multi-object interactions. Consequently, prior methods address only subsets of the three challenges, leaving the full problem unresolved.

We propose \textbf{MoDex}, a data-efficient framework for universal multi-object dexterous grasping. Our core insight is that a single reference trajectory already encodes rich interaction priors, allowing diverse object instances to be handled by learning compact residual corrections rather than a full action sequence. This reduces the long-horizon problem to a single-step Markov Decision Process. We further observe that the optimal strategy is layout-dependent: proximal objects favor efficient single-shot enclosure, whereas separated objects require sequential interaction. Accordingly, we introduce a spatially adaptive dual-strategy mechanism that handles both regimes under the same residual formulation. Finally, we train a vision-based policy solely from randomized simulation rollouts, enabling zero-shot real-world deployment.

Experiments show that MoDex substantially outperforms baselines in dual-object proximal layouts, achieving 78.73\% success on seen object pairs and 61.24\% on unseen objects, which further improve to 80.74\% and 63.69\% with distillation. In separated layouts, MoDex reaches 92.76\% success on seen object pairs and 39.85\% on unseen objects. We also validate its effectiveness across different robot embodiments and a larger object set. In real-world tests, our vision-based policy achieves a 55\% success rate on 20 unseen objects in proximal layouts and 45.5\% in separated layouts, without using any real-world training data.

Our contributions are threefold. First, we show that multi-object dexterous grasping can be learned data-efficiently by adapting a single demonstration through a single-step residual policy. Second, we offer a unified solution to diverse object layouts, seamlessly integrating single-shot and multi-stage strategies within one spatially adaptive framework. Third, our method generalizes across unseen geometries, layouts, embodiments, and achieves zero-shot sim-to-real transfer by a vision-based policy trained exclusively on simulation data.

\section{Related Work}
\noindent\textbf{Multi-object Grasping.} Multi-object grasping aims to enclose and stabilize multiple objects in one hand, introducing spatial competition and force constraints beyond single-object grasping~\citep{agboh2022multi,harada1998kinematics,sun2022multi,yamada2015static,yao2023exploiting}. Recent works address this via grasp generation: MultiGrasp~\citep{li2024grasp} identifies hand configurations through differentiable force closure optimization~\citep{liu2021synthesizing} or diffusion models~\citep{ho2020denoising}, then executes the grasp with a goal-conditioned RL policy; SeqGrasp~\citep{lu2025grasping} partitions the hand's degrees of freedom for sequential synthesis. SeqMultiGrasp~\citep{he2025sequential} likewise addresses sequential multi-object grasping through grasp synthesis and generation. In contrast, we bypass expensive optimization and learn a universal policy via demonstration-guided RL. Our spatially adaptive dual-strategy mechanism further enables the first unified framework handling both proximal and separated layouts.

\noindent\textbf{Reinforcement Learning for Dexterous Manipulation.} Deep RL enables high-dimensional continuous control through environment interaction~\citep{mnih2015human,DBLP:journals/corr/LillicrapHPHETS15,schulman2017proximal,haarnoja2018soft}. To master dexterous contact dynamics, recent methods structure learning for scalability and sample efficiency. To scale across thousands of objects, UniDexGrasp~\citep{xu2023unidexgrasp} pioneered task decoupling and geometry-aware curriculum distillation, while UniDexGrasp++~\citep{wan2023unidexgrasp++} later advanced this approach with iterative specialist–generalist training. Residual learning frameworks~\citep{huang2025efficient,su2026rfs,DBLP:conf/rss/HuMS24} improve sample efficiency by learning corrections on top of pre-trained policies or priors. DemoGrasp~\citep{yuan2025demograsp} adapts an expert demonstration as a single-step MDP, yielding a universal grasping policy. However, these learning frameworks are designed for single-object manipulation. Inspired by their success, we learn residual corrections to one demonstration through single-step reinforcement learning. 

\noindent\textbf{Imitation Learning for Dexterous Manipulation.} Imitation learning~\citep{argall2009survey,hussein2017imitation,osa2018algorithmic} avoids complex reward design by learning from expert trajectories. Foundational behavioral cloning~\citep{pomerleau1988alvinn} and inverse RL~\citep{ng2000algorithms,ziebart2008maximum} struggle to scale to dexterous hands due to data scarcity and high-DoF control. Recent works bridge this gap by collecting large-scale human demonstrations: DexMV~\citep{qin2022dexmv} and AnyTeleop~\citep{qin2023anyteleop} use visual retargeting and teleoperation. Diffusion policies~\citep{chi2025diffusion,DBLP:conf/rss/ZeZZHWX24} are employed to model the intricate, high-DoF action spaces~\citep{xu2025dexumi,DBLP:conf/rss/SiZTK24,liang2025dexhanddiff}. However, massive data collection remains expensive. In contrast, our method distills necessary control priors from only a single demonstration, removing the need for exhaustive data and complex morphological mapping.

\section{Method}
We present MoDex, a framework that learns a universal multi-object dexterous grasping policy from a single demonstration. Following the problem formulation (Sec.~\ref{subsec:problem}), we introduce our single-step residual policy (Sec.~\ref{subsec:alg}). A spatially adaptive dual strategy mechanism is further proposed to explicitly address different object layouts, and we utilize a specialist-generalist distillation pipeline to improve performance in proximal layouts (Sec.~\ref{subsec:dual}). Finally, we obtain a vision-based policy for real-world deployment by fine-tuning a vision-language-action model using exclusively simulation data (Sec.~\ref{subsec:sim2real}).

\begin{figure}[t]
    \centering
    \includegraphics[width=\linewidth]{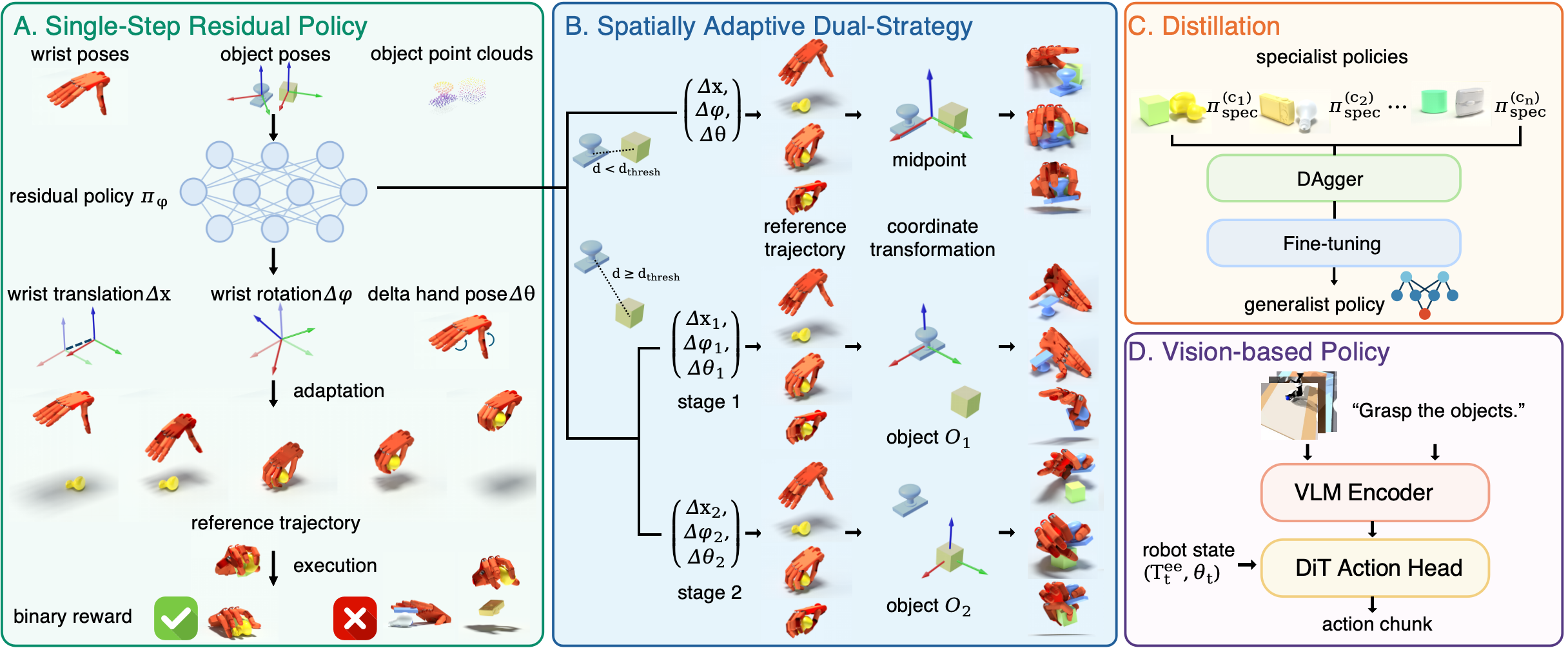}
    \vspace{-3pt}
    \caption{Overview of MoDex. (A) A single-step residual policy adapts a reference trajectory with residual corrections. (B) A spatially adaptive dual-strategy handles proximal layouts with single-shot grasping and separated layouts with multi-stage grasping. (C) Specialists are distilled into a generalist via DAgger and fine-tuning. (D) The architecture of our vision-based policy.}
    \label{fig:method}
    \vspace{-6pt}
\end{figure}

\subsection{Problem Formulation}
\label{subsec:problem}
We consider table-top multi-object grasping where a dexterous hand lifts \(N=2\) objects within horizon \(T\). Let \(d_{\min}\) be the minimal pairwise planar distance between the objects' projected 2D positions on the table. We call a configuration \textit{proximal} if \(d_{\min} < d_{\text{thresh}}\) and \textit{separated} otherwise. The observation \(\mathbf{o}_t\) contains the robot proprioception (6D wrist pose \(\mathbf{T}^{ee}_t\) and joint angles \(\boldsymbol{\theta}_t\)), object poses \(\{\mathbf{T}^o_{t,i}\}_{i=1}^N\) and object point clouds \(\{\mathcal{P}_i\}_{i=1}^N\). Action \(\mathbf{a}_t\) includes target wrist pose and joint angles. The objective is a policy \(\pi(\mathbf{a}_t \mid \mathbf{o}_t)\) maximizing the expected discounted return while generalizing across object geometries and layouts.

\subsection{Single-Step Residual Policy}
\label{subsec:alg}
Grasping tasks that involve different objects share a common three-phase structure: reaching, grasping, and lifting. The reaching phase can typically be handled by motion planning, while the grasping and lifting phases demand precise interaction with the objects. Following~\citep{yuan2025demograsp}, we capture the pattern of the latter two phases using a single reference trajectory \(\tau^{\text{ref}} = \{(\mathbf{x}_{\text{obj},t}^{\text{ref}}, \mathbf{q}_{\text{obj},t}^{\text{ref}}, \boldsymbol{\theta}_t^{\text{ref}})\}_{t=0}^{T_{\text{ref}}}\) recorded while grasping an arbitrary object and shared as a prior across all training scenes. Here \(\mathbf{x}_{\text{obj},t}^{\text{ref}} \in \mathbb{R}^3\) and \(\mathbf{q}_{\text{obj},t}^{\text{ref}} \in \mathbb{S}^3\) denote the wrist position and orientation, both expressed in the initial object frame. The remaining component \(\boldsymbol{\theta}_t^{\text{ref}}\) specifies the hand joint configuration. Let \(T_{\text{lift}}\) be the timestep the lifting phase commences. As shown in part A of Figure~\ref{fig:method}, rather than learning the full action sequence \(\{\mathbf{a_t}\}_{t=0}^T\) over the entire episode, our policy outputs a compact residual action vector \(\mathbf{r} = (\Delta\mathbf{x}, \Delta\boldsymbol{\phi}, \Delta\boldsymbol{\theta})\), which is then rescaled component-wise by fixed ranges. The scaled quantities produce a wrist translation \(\Delta\mathbf{x'}\), a wrist rotation (Euler angles) \(\Delta\boldsymbol{\phi'}\), and a hand joint angle offset that is added to the reference grasp pose \(\boldsymbol{\theta}_{T_{\text{lift}}}^{\text{ref}}\) to obtain the final hand configuration \(\tilde{\boldsymbol{\theta}}_{\text{final}}\). The adapted wrist trajectory is obtained by applying an \(\text{SE}(3)\) transformation to the grasping segment; for the lifting part, the relative displacement from the reference lifting motion is replayed starting from the new pre-lifting pose:
\begin{equation*}
\begin{aligned}
\tilde{\mathbf{q}}_{\text{obj},t}
&= \operatorname{Quat}(\Delta\boldsymbol{\phi}') \otimes \mathbf{q}_{\text{obj},t}^{\text{ref}}, \\
\tilde{\mathbf{x}}_{\text{obj},t}
&=
\begin{cases}
\operatorname{Rot}(\Delta\boldsymbol{\phi}') \mathbf{x}_{\text{obj},t}^{\text{ref}} + \Delta\mathbf{x}', 
& t \le T_{\text{lift}}, \\[4pt]
\tilde{\mathbf{x}}_{\text{obj},T_{\text{lift}}}
+
\left(
\mathbf{x}_{\text{obj},t}^{\text{ref}}
-
\mathbf{x}_{\text{obj},T_{\text{lift}}}^{\text{ref}}
\right),
& t > T_{\text{lift}} .
\end{cases}
\end{aligned}
\end{equation*}
The hand joint trajectory before timestep \(T_{\text{lift}}\) is generated by an affine rescaling that preserves the temporal profile of the reference grasping motion while setting its terminal value to \(\tilde{\boldsymbol{\theta}}_{\text{final}}\); the joints are then held constant throughout the lifting phase:
\begin{equation*}
\tilde{\boldsymbol{\theta}}_t =
\begin{cases}
\boldsymbol{\theta}_0^{\text{ref}} + \bigl(\boldsymbol{\theta}_t^{\text{ref}} - \boldsymbol{\theta}_0^{\text{ref}}\bigr) \odot \dfrac{\tilde{\boldsymbol{\theta}}_{\text{final}} - \boldsymbol{\theta}_0^{\text{ref}}}{\boldsymbol{\theta}_{T_{\text{lift}}}^{\text{ref}} - \boldsymbol{\theta}_0^{\text{ref}}}, & t \le T_{\text{lift}}, \\[12pt]
\tilde{\boldsymbol{\theta}}_{\text{final}}, & t > T_{\text{lift}}.
\end{cases}
\end{equation*}
Finally, we obtain a modified trajectory \(\tau^{\text{new}} = \{(\tilde{\mathbf{x}}_{\text{obj},t}, \tilde{\mathbf{q}}_{\text{obj},t}, \tilde{\boldsymbol{\theta}}_t)\}_{t=0}^{T_{\text{ref}}}\). Then we transform the wrist poses from the new object frame to the world frame and get \((\tilde{\mathbf{x}}_t, \tilde{\mathbf{q}}_t)\). How the object frame is defined in multi-object scenes is discussed later in Sec.~\ref{subsec:dual}. The corresponding arm joint targets are computed via inverse kinematics from the adapted wrist poses \((\tilde{\mathbf{x}}_t, \tilde{\mathbf{q}}_t)\). This new trajectory can be directly executed open-loop. In this way, the action space is reduced from a high-dimensional per-timestep sequence to the compact low-dimensional bounded residual \(\mathbf{r}\). 

We formulate the resulting task as a single-step decision problem. The policy receives an initial observation \(\mathbf{o}_0\) comprising robot proprioception, object poses and object point clouds, then outputs the residual vector \(\mathbf{r}\). 
Given \(\mathbf{r}\), a new trajectory is constructed via the rules above and executed open-loop until termination, after which a sparse binary reward is obtained. Specifically, let \(n_{\text{lift}}\) be the number of objects lifted at the end of the episode, we define the terminal reward as:
\[
R(\mathbf{o}_0,\mathbf{r}) =
\mathbb{1}_{\text{no collision}} \cdot \mathbb{1}_{n_{\text{lift}} = 2},
\]
which equals \(1\) only when both objects are lifted without hand-table collision. Since only a single decision is made at the start of each episode, the accumulated return reduces to this terminal reward. The objective is to optimize the policy parameters \(\psi\) by maximizing
\[
J(\psi)
=
\mathbb{E}_{\mathbf{o}_0,\, \mathbf{r} \sim \pi_{\psi}(\cdot \mid \mathbf{o}_0)}
\left[
R(\mathbf{o}_0,\mathbf{r})
\right],
\]
where \(\pi_{\psi}\) denotes the residual policy. 
We optimize this objective using Proximal Policy Optimization~\citep{schulman2017proximal}.

\subsection{Spatially Adaptive Dual-Strategy}
\label{subsec:dual}
The optimal grasping strategy depends strongly on the spatial arrangement of the objects. In proximal layouts, where the objects are placed close to each other, a single-shot grasp that envelops both objects tends to be most effective. In separated layouts, where they are far apart, grasping them sequentially with different finger groups yields better results. We therefore design a spatially adaptive dual-strategy mechanism that switches based on $d_{\min}$. If $d_{\min} < d_{\text{thresh}}$ it invokes the single-shot strategy, otherwise the multi-stage strategy. Both strategies are built upon the same single-step residual formulation introduced in Sec.~\ref{subsec:alg}, but they differ in the structure of the residual action and the execution protocol, which is visualized in part B of Figure~\ref{fig:method}.

\textbf{Single-shot Strategy.} For proximal layouts, the residual has the same form as in Sec.~\ref{subsec:alg}. All fingers close simultaneously to grasp both objects in one attempt. \(\tau^{\text{new}}\) and all observations are expressed in a coordinate frame centered at the midpoint of the two objects. Furthermore, as illustrated in part C of Figure~\ref{fig:method}, we improve performance by first training specialists per object combination and then distilling them into a generalist via DAgger~\citep{ross2011reduction}, followed by a fine-tuning stage. Each specialist $\pi_{\text{spec}}^{(c)}$ is trained independently on a pair of objects $c$, using the same policy in Sec.~\ref{subsec:alg}. A mixed strategy then selects the specialist with probability $\beta_k$, which is linearly annealed from 1 to 0, and selects the current generalist otherwise, to acquire initial observation $\mathbf{o}_0$. For each collected $\mathbf{o}_0$, the corresponding specialist provides a target residual $\bar{\mathbf{r}}^{(c)}$, and the generalist $\pi_\psi$ minimizes the following loss function:
\[
\mathcal{L}_{\text{BC}}
=
\mathbb{E}_{\mathbf{o}_0,c}
\left[
\left\|
\pi_\psi(\mathbf{o}_0)
-
\operatorname{atanh}\left(\bar{\mathbf{r}}^{(c)}\right)
\right\|_2^2
\right].
\]
After distillation, the generalist successfully mimics the specialists' behaviors but its performance on out-of-distribution scenarios is not good enough. We thus reinitialize a network from the generalist and conduct a task-level fine-tuning as in Sec.~\ref{subsec:alg}.

\textbf{Multi-stage Strategy.} For separated layouts, we split the grasping process into two sequential stages, each handled by a dedicated finger group. The residual is extended to two independent sets: $(\Delta\mathbf{x}_{1}, \Delta\boldsymbol{\phi}_{1}, \Delta\boldsymbol{\theta}_{1})$ for the first object and $(\Delta\mathbf{x}_{2}, \Delta\boldsymbol{\phi}_{2}, \Delta\boldsymbol{\theta}_{2})$ for the second. \(\Delta\boldsymbol{\theta}_{1}\) contains the correction to the actions of thumb and index fingers, and the joints of the other fingers are frozen to the initial state in the first stage. The adapted \(\tau^{\text{new}}_1\) is transformed from the object frame of \(O_1\) to the world frame, and then executed to grasp object $O_1$. Once $O_1$ is lifted above a height threshold, the grip is frozen. In the second stage, \(\Delta\boldsymbol{\theta}_{2}\) controls the remaining three fingers. \(\tau^{\text{new}}_2\) is transformed from the object frame of \(O_2\) to the world frame to grasp $O_2$ with the thumb and index fingers locked. Finally both objects are lifted together.

\subsection{Vision-based Policy Learning}
\label{subsec:sim2real}
We distill the state-based policy into a vision-based policy that takes only RGB images and robot proprioceptive states as input. The model architecture is depicted in part D of Figure~\ref{fig:method}. Specifically, we roll out the trained state-based policy in a visually randomized simulation across a wide variety of objects and initial configurations, retaining only successful episodes. At each timestep, we record RGB images $\{\mathbf{I}_{t,i}\}_{i=0}^{n}$, the robot proprioceptive state \((\mathbf{T}^{\text{ee}}_t, \boldsymbol{\theta}_t)\), as well as the corresponding low-level command $\mathbf{a}_t$ issued to the controller. We subsequently fine-tune a vision-language-action model~\citep{bjorck2025gr00t} on the collected dataset. Given the current observation \((\{\mathbf{I}_{t,i}\}_{i=0}^n, \mathbf{T}^{\text{ee}}_t, \boldsymbol{\theta}_t)\), the model predicts a sequence of future actions $\{\mathbf{a}_{t+i}\}_{i=0}^{H-1}$ using a flow-matching~\citep{lipman2022flow} action head. The learned vision-based policy can be directly deployed on the real robot without requiring additional training on any real-world demonstration.


\section{Experiments}
\label{sec:result}
\subsection{Simulation Experimental Settings}
\label{sec:sim_exp_settings}
We use the IsaacGym~\citep{makoviychuk2021isaac} simulator. Following~\citep{li2024grasp}, unless specified, the experiments are conducted with a Shadow Hand on 11 objects from YCB~\citep{calli2017yale} and ContactDB~\citep{brahmbhatt2019contactdb}, with diverse pose randomization based on 115 pre-computed stable resting rotations. We train on 28 combinations formed by 8 objects. Proximal layouts are distinguished from separated layouts; the latter imposes a minimum pairwise distance $d_{\text{thresh}}\in\{0.15,0.20,0.25\}\,\text{m}$. Evaluation covers four conditions: \textit{Training} (seen pairs and poses), \textit{Unseen Comb.} (unseen combinations of seen objects), \textit{Unseen Place.} (seen pairs, unseen poses), and \textit{Unseen Geom.} (unseen objects). A trial succeeds if both objects are lifted 0.1\,m higher without hand-table collision. For proximal baselines, we compare with \textit{MultiGrasp-Syn}~\citep{li2024grasp}, \textit{MultiGrasp-Gen}~\citep{li2024grasp}, and adaptations of \textit{UniDexGrasp++}~\citep{wan2023unidexgrasp++} and \textit{ResDex}~\citep{huang2025efficient}. For separated layouts, we construct two baselines \textit{MoDex-Single} (our single-shot strategy) and \textit{Gather-Single} (push objects together and then grasp). MoDex is evaluated using 8,000-iteration checkpoints, while all baselines are trained until convergence.

We compare with SeqMultiGrasp-SG~\citep{he2025sequential}, adapted to IsaacGym on Panda+Allegro, over 1,600 matched separated-layout trials (64 known object pairs, 25 layouts each). Terminal success requires both objects above 10\,cm relative to the initial pose.

\subsection{Main Results}
\textbf{Performance on Proximal Layouts.} Table~\ref{tab:main_results} summarizes the results for proximal layouts. Our method consistently outperforms baselines across all settings, achieving an 78.73\% success rate on the training set, which is 21.43 percentage points higher than the strongest baseline. This performance gain may stem from our effective use of the reference trajectory, which shortens the temporal horizon of policy learning and constrains the action space through residual correction. On unseen objects with distinct geometries, our method maintains a 61.24\% success rate, indicating a certain degree of generalization ability. Moreover, the specialist-generalist distillation pipeline further improves performance over the non-distilled variant, suggesting that specialist policies provide useful supervision for learning a stronger generalist policy.

\begin{table}[!t]
\centering
\begin{minipage}{\linewidth}
\setlength{\abovecaptionskip}{0pt}
\setlength{\belowcaptionskip}{0pt}
\caption{\textbf{Main simulation results.} Success rates (\%) under proximal and separated layouts across four evaluation settings. Separated-layout results are averaged over $d_{\mathrm{thresh}} \in \{0.15, 0.20, 0.25\}$ m.}
\label{tab:main_results}

\resizebox{\linewidth}{!}{
\begin{tabular}{lcccc}
\toprule
Method & Training & Unseen Comb. & Unseen Place. & Unseen Geom. \\
\midrule

\multicolumn{5}{l}{\textit{Proximal layout}} \\
MultiGrasp-Syn~\citep{li2024grasp}$^\dagger$ & -- & 26.73 & 68.34 & 22.55 \\
MultiGrasp-Gen~\citep{li2024grasp}$^\dagger$ & -- & 23.32 & 40.20 & 15.65 \\
UniDexGrasp++~\citep{wan2023unidexgrasp++} & 57.30 & 56.08 & 56.61 & 28.21 \\
ResDex~\citep{huang2025efficient} & 53.85 & 53.98 & 55.76 & 27.54 \\
\rowcolor{gray!15}
MoDex (Ours) & \underline{78.73} & \underline{78.55} & \underline{82.00} & \underline{61.24} \\
\rowcolor{gray!25}
MoDex w/ distillation (Ours) & \textbf{80.74} & \textbf{80.79} & \textbf{84.41} & \textbf{63.69} \\
\midrule

\multicolumn{5}{l}{\textit{Separated layout}} \\
MoDex-Single & \underline{18.89} & \underline{18.18} & 17.60 & 9.84 \\
Gather-Single & 17.73 & 17.27 & \underline{18.11} & \underline{11.49} \\
\rowcolor{gray!15}
MoDex (Ours) & \textbf{92.76} & \textbf{92.27} & \textbf{84.21} & \textbf{39.85} \\
\bottomrule
\end{tabular}
}
\vspace{1pt}
{\footnotesize
$^\dagger$ Results are taken from~\citep{li2024grasp}, as the evaluation conditions are the same.
}
\end{minipage}
\vspace{-2.0em}
\end{table}

\begin{wraptable}{R}{0.36\textwidth}
    \vspace{-1.5em}
    \centering
    \small
    \setlength{\abovecaptionskip}{2pt}
    \setlength{\belowcaptionskip}{3pt}
    \setlength{\tabcolsep}{3pt}
    \caption{\textbf{Sequential grasping.} Terminal success rates (\%) in our adapted setting.}
    \label{tab:seqmultigrasp_comparison}
    \begin{tabular}{@{}l@{\hspace{12pt}}r@{}}
    \toprule
    Method & Success \\
    \midrule
    SeqMultiGrasp-SG & 75.00 \\
    MoDex (Ours) & \textbf{99.75} \\
    \bottomrule
    \end{tabular}
    \vspace{-1em}
\end{wraptable}

\textbf{Performance on Separated Layouts.} 
Handling spatially distributed objects requires a fundamentally different strategy. As shown in Table~\ref{tab:main_results}, directly applying the single-shot strategy \textit{MoDex-Single} leads to poor performance, as the hand can't cover both objects simultaneously. The heuristic baseline \textit{Gather-Single} also performs poorly because pushing irregular objects often results in unpredictable collisions or toppling. The success rates of these two baselines drop to $0$ when $d_{\text{thresh}} > 0.15 \text{m}$. In contrast, our spatially adaptive strategy effectively tackles this challenge, achieving a 92.76\% success rate on the training set and 39.85\% on novel object geometries. In the additional Panda+Allegro comparison (Table~\ref{tab:seqmultigrasp_comparison}), MoDex achieves 99.75\% terminal success versus 75.00\% for SeqMultiGrasp-SG.

\begin{figure}[t]
    \centering
    \setlength{\abovecaptionskip}{-1em}
    \setlength{\belowcaptionskip}{0pt}
    \includegraphics[width=\linewidth, trim=9pt 9pt 7pt 5pt, clip]{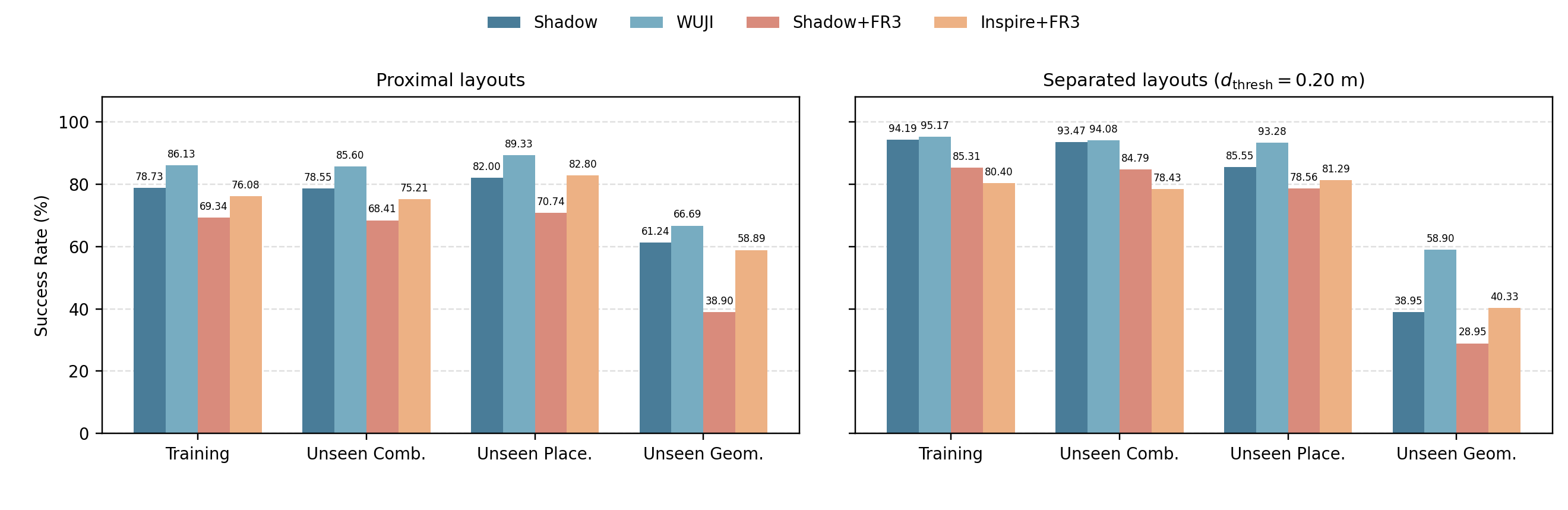}
    \caption{\textbf{Cross-embodiment results.} Success rates (\%) of our method across different hand morphologies and arm-hand systems under proximal and separated layouts.}
    \label{fig:barplot_1}
\end{figure}

\textbf{Cross-Embodiment Experiment.} We conduct experiments on alternative hand morphologies and full arm-hand systems. The results are shown in Figure~\ref{fig:barplot_1}. For proximal layouts, the floating WUJI Hand yields even higher success rates (86.13\% on the training set, 66.69\% on unseen objects). When mounted on a Franka Research 3 (FR3) arm, the Shadow+FR3 system retains competitive performance (69.34\% and 38.90\%) despite the added complexity of inverse kinematics. A similar trend is observed in separated layouts, where the WUJI Hand reaches 95.17\% success rate on seen configurations. The Inspire+FR3 setup also achieves good results despite its underactuated fingers which are less suited for sequential grasping.

\begin{wrapfigure}{r}{0.35\textwidth} 
    \setlength{\abovecaptionskip}{-1em}
    \vspace{-2em}
    \centering
    \includegraphics[width=\linewidth, trim=5pt 10pt 5pt 5pt, clip]{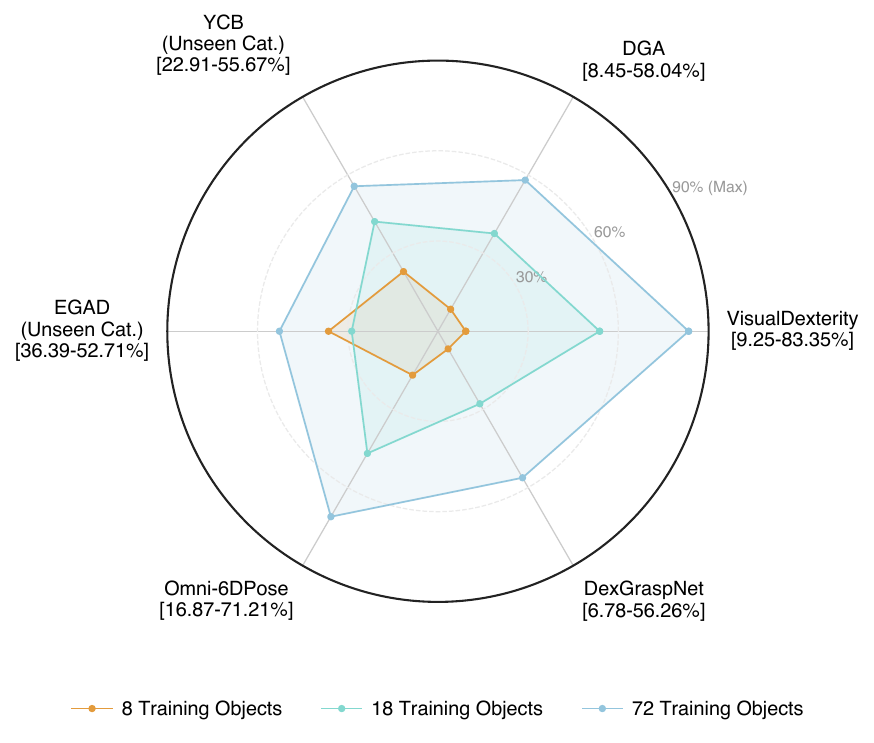}
    \caption{Scaling leads to better out-of-distribution performance.}
    \label{fig:scaling}
    \vspace{-2em}
\end{wrapfigure}

\textbf{Scaling to Diverse Objects.}
Previous research on multi-object grasping has typically evaluated on a narrow set of objects, leaving method scalability unexplored. We expand the training set to 18 objects and 72 objects respectively, selected from YCB~\citep{calli2017yale}, ContactDB~\citep{brahmbhatt2019contactdb} and EGAD~\citep{DBLP:journals/ral/MorrisonCL20}. We test on unseen objects selected from six datasets: YCB~\citep{calli2017yale}, DGA~\citep{DBLP:conf/cvpr/ZhongJYM25}, VisualDexterity~\citep{DBLP:journals/scirobotics/ChenTWKAA23}, DexGraspNet~\citep{wang2022dexgraspnet}, Omni-6DPose~\citep{DBLP:conf/eccv/ZhangHPWHCZD24} and EGAD~\citep{DBLP:journals/ral/MorrisonCL20}. Results in Figure~\ref{fig:scaling} indicate that generalization improves with more training objects.

\WFclear
\subsection{Ablation Studies}
\label{subsec:ablation}

\begin{table}[H]
\vspace{-4pt}
\centering
\begin{minipage}{\linewidth}
\caption{\textbf{Component ablations.} Success rates (\%) of our full method and its ablated variants.}
\label{tab:ablation}

\resizebox{\linewidth}{!}{
\begin{tabular}{lcccc}
\toprule
Ablation variants & Training & Unseen Comb. & Unseen Place. & Unseen Geom. \\
\midrule
w/o Residual Correction & 0.42 & 0.65 & 0.49 & 0.80 \\
w/o Single Reference & 3.54 & 3.76 & 4.83 & 5.62 \\
w/o Midpoint Coord. Trans. & 74.69 & 74.21 & 79.85 & 59.89 \\
\rowcolor{gray!15}
MoDex (Ours) & \textbf{78.73} & \textbf{78.55} & \textbf{82.00} & \textbf{61.24} \\
\bottomrule
\end{tabular}
}
\end{minipage}
\vspace{-4pt}
\end{table}

\textbf{Component Ablations.}
Table~\ref{tab:ablation} evaluates three key components in the proximal single-shot setting. Using a randomly selected object frame instead of the midpoint frame consistently reduces performance, suggesting that expressing observations and adapted trajectories in the object-midpoint frame provides a useful normalization. Replacing the single reference with MultiGrasp~\citep{li2024grasp} pre-grasp configurations causes a severe drop in success rates, indicating that the reference trajectory provides an execution-level motion prior rather than only a static grasp pose. Directly replaying the reference trajectory without residual correction almost completely fails, confirming that the learned residual is essential for adapting one demonstration to different object geometries and placements.

\textbf{Reference Trajectory Sensitivity.}
We compare a top-grasp reference recorded on a duck with a side-grasp reference recorded on a cube, using the floating Shadow Hand. As shown in Table~\ref{tab:reference_sensitivity}, changing the reference alters training, unseen-combination, and unseen-placement success by at most 2.45 percentage points. Unseen-geometry success is similar in separated layouts (38.95\% versus 39.06\%), but decreases from 61.24\% to 52.63\% in proximal layouts. Thus, both references support effective learning, while reference choice can affect geometric generalization.

\begin{table}[H]
\vspace{-4pt}
\centering
\small
\setlength{\tabcolsep}{3.5pt}
\caption{\textbf{Reference trajectory sensitivity.} Success rates (\%) with the floating Shadow Hand. Separated layouts use $d_{\mathrm{thresh}}=0.20$\,m.}
\label{tab:reference_sensitivity}
\begin{tabularx}{\linewidth}{@{}llYYYY@{}}
\toprule
Reference & Layout & Training & Unseen Comb. & Unseen Place. & Unseen Geom. \\
\midrule
Duck / top & Proximal & 78.73 & 78.55 & 82.00 & 61.24 \\
Cube / side & Proximal & 80.72 & 79.67 & 81.65 & 52.63 \\
\midrule
Duck / top & Separated & 94.19 & 93.47 & 85.55 & 38.95 \\
Cube / side & Separated & 92.30 & 91.38 & 88.00 & 39.06 \\
\bottomrule
\end{tabularx}
\vspace{-4pt}
\end{table}

\textbf{Reward Design.} We compare two reward designs in Table~\ref{tab:reward_ablation}. The binary reward assigns a reward of $1$ only when all objects are lifted, whereas the curriculum reward assigns a reward equal to the number of lifted objects. Overall, the binary reward achieves higher performance on both training configurations (+23.79\%) and unseen configurations (+19.04\%), avoiding severe failures observed with curriculum reward in some settings. One possible explanation is that curriculum reward may provide a misaligned intermediate signal: lifting only one object can be rewarded even when the resulting behaviour does not lead to successful grasping of two objects.

\begin{table}[H]
\vspace{-4pt}
\centering
\caption{\textbf{Reward design ablation.} Success rates (\%) for binary and curriculum rewards across different embodiments and layouts. Unseen Avg. averages over unseen object combinations, placements, and geometries, and $\Delta$ denotes Binary $-$ Curriculum.}
\label{tab:reward_ablation}
\small
\setlength{\tabcolsep}{2.8pt}
\renewcommand{\arraystretch}{1.02}
\newcolumntype{Y}{>{\centering\arraybackslash}X}
\begin{tabularx}{\linewidth}{@{}llcYYYYYY@{}}
\toprule
\multirow{2}{*}{Embodiment}
& \multirow{2}{*}{Layout}
& \multirow{2}{*}{$d_{\mathrm{thresh}}$}
& \multicolumn{3}{c}{Training}
& \multicolumn{3}{c}{Unseen Avg.} \\
\cmidrule(lr){4-6} \cmidrule(l){7-9}
& & & Binary & Curr. & $\Delta$ & Binary & Curr. & $\Delta$ \\
\midrule
Shadow & Proximal  & --   & 78.73 & \textbf{90.09} & $-11.36$ & 73.93 & \textbf{77.97} & $-4.04$ \\
Shadow & Separated & 0.15 & \textbf{90.50} & 76.69 & $+13.81$ & \textbf{72.02} & 60.61 & $+11.41$ \\
Shadow & Separated & 0.20 & \textbf{94.19} & 93.51 & $+0.68$  & \textbf{72.65} & 71.98 & $+0.67$ \\
Shadow & Separated & 0.25 & 93.58 & \textbf{94.00} & $-0.42$  & 71.66 & \textbf{72.37} & $-0.71$ \\
\midrule
WUJI & Separated & 0.20 & 95.17 & \textbf{97.02} & $-1.86$ & 82.09 & \textbf{86.84} & $-4.75$ \\
Shadow+FR3 & Separated & 0.20 & \textbf{85.31} & 0.01 & $+85.30$ & \textbf{64.10} & 0.09 & $+64.01$ \\
Inspire+FR3 & Separated & 0.20 & \textbf{80.40} & 0.00 & $+80.40$ & \textbf{66.68} & 0.00 & $+66.68$ \\
\midrule
\multicolumn{3}{@{}l}{\textit{Overall avg.}}
& \textbf{88.27} & 64.47 & $+23.79$
& \textbf{71.88} & 52.84 & $+19.04$ \\
\bottomrule
\end{tabularx}
\vspace{-4pt}
\end{table}

\textbf{Distillation Method.}
Table~\ref{tab:ablate_distillation} compares different ways of distilling specialist policies into a generalist. \textit{BC-MLP} trains a deterministic MLP policy offline using collected specialist rollouts. \textit{BC-MoG} uses the same offline data but replaces the deterministic action head with a mixture-of-Gaussians action head~\citep{bishop1994mixture}. DAgger~\citep{ross2011reduction} improves over offline BC by collecting supervision under the mixed specialist-generalist rollout distribution, yet its performance on out-of-distribution scenarios remains limited. Our fine-tuning stage directly optimizes task-level success after distillation, leading to substantially better generalization.

\begin{table}[H]
\vspace{-4pt}
\centering
\begin{minipage}{\linewidth}
\caption{\textbf{Distillation method ablation.} Success rates (\%) of different distillation schemes.}
\label{tab:ablate_distillation}

\small
\renewcommand{\arraystretch}{1.02}
\newcolumntype{Y}{>{\centering\arraybackslash}X}

\begin{tabularx}{\linewidth}{>{\raggedright\arraybackslash}XYYYY}
\toprule
Method & Training & Unseen Comb. & Unseen Place. & Unseen Geom. \\
\midrule
BC-MLP & 44.73 & 28.70 & 48.47 & 24.55 \\
BC-MoG & 75.27 & 34.42 & 71.51 & 22.17 \\
DAgger & 80.60 & 37.80 & 78.80 & 28.50 \\
\rowcolor{gray!25}
Ours & \textbf{80.74} & \textbf{80.79} & \textbf{84.41} & \textbf{63.69} \\
\bottomrule
\end{tabularx}
\end{minipage}
\vspace{-4pt}
\end{table}

\subsection{Real Experiments}
\label{sec:real_experiments}
\begin{figure}[H]
    \centering
    \includegraphics[width=\linewidth]{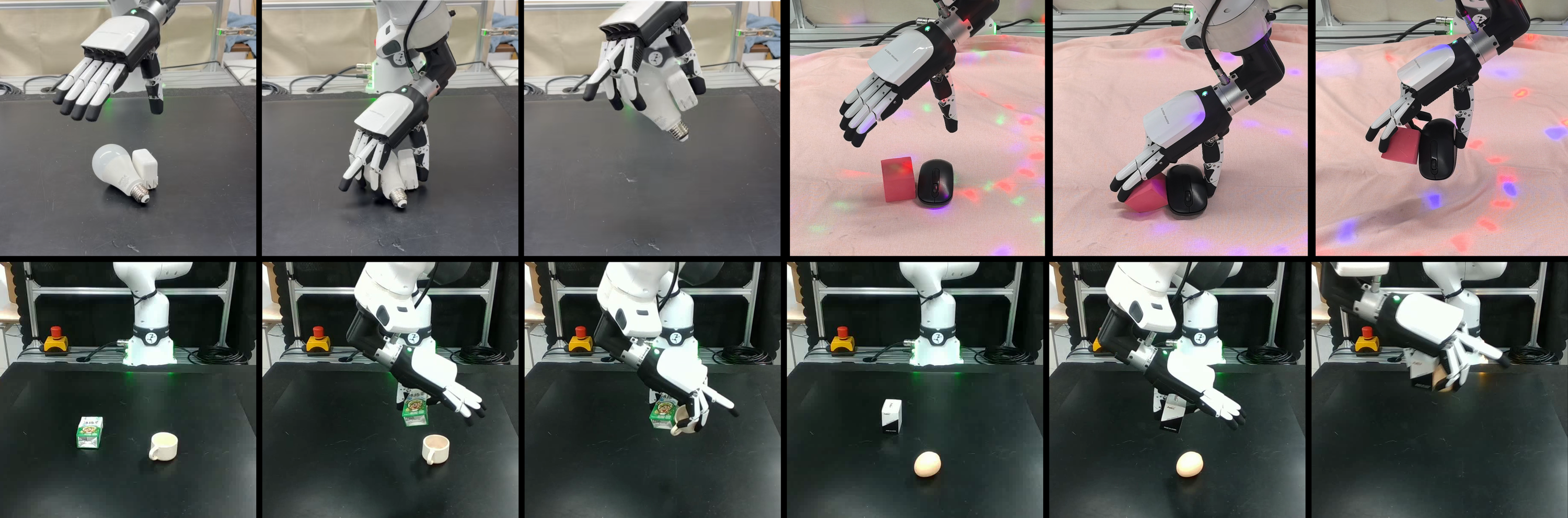}
    \vspace{-3pt}
    \caption{\textbf{Real-world experiments.} Successful zero-shot executions in proximal (top) and separated (bottom) layouts. Each three-frame sequence shows one trial progressing from left to right.}
    \label{fig:real_experiment}
    \vspace{-6pt}
\end{figure}

We evaluate the vision-based policy on a real robot with a 6-DoF Inspire Hand mounted on a 7-DoF Franka Research 3 arm. The test set contains 20 unseen real-world objects across 10 combinations, with 20 randomized trials per combination. Our policy achieves a 55\% success rate in proximal layouts and a 45.5\% success rate in separated layouts, demonstrating zero-shot transfer to real-world objects. As shown in Figure~\ref{fig:real_experiment}, successful trials cover both proximal and separated layouts, as well as diverse backgrounds and lighting conditions, indicating a degree of robustness to visual variations. More details are provided in the Appendix.

\section{Limitations}
Our method has several limitations. First, experiments are conducted only on dual-object settings. Second, although scaling potential is demonstrated, our approach has not been validated on a massive-scale object dataset. Third, sim-to-real transfer exhibits a noticeable performance gap. Future work will focus on extending the policy to more objects and incorporating multi-modal feedback to improve real-world reliability.

\section{Conclusion}
\label{sec:conclusion}
We propose MoDex, a data-efficient framework that learns universal multi-object dexterous grasping. Our method formulates the task as a single-step residual policy learning problem and introduces a spatially adaptive dual-strategy to handle diverse object layouts. Simulation experiments demonstrate substantial improvements over prior methods, effective transfer across different embodiments, and scalability, while real-world experiments show zero-shot sim-to-real transfer on unseen objects.


\acknowledgments{This work was supported by NSFC in part under Grant 62450001 and 62476008. The authors would like to thank the anonymous reviewers for their valuable comments and advice. We thank Yang Tian for his helpful writing suggestions and for treating the first author to many delicious meals during the final push to finish this paper.}


\clearpage
\bibliography{example}  

@article{li2024grasp,
  title={Grasp multiple objects with one hand},
  author={Li, Yuyang and Liu, Bo and Geng, Yiran and Li, Puhao and Yang, Yaodong and Zhu, Yixin and Liu, Tengyu and Huang, Siyuan},
  journal={IEEE Robotics and Automation Letters},
  volume={9},
  number={5},
  pages={4027--4034},
  year={2024},
  publisher={IEEE}
}

@article{lu2025grasping,
  title={Grasping a Handful: Sequential Multi-Object Dexterous Grasp Generation},
  author={Lu, Haofei and Dong, Yifei and Weng, Zehang and Pokorny, Florian and Lundell, Jens and Kragic, Danica},
  journal={IEEE Robotics and Automation Letters},
  year={2025},
  publisher={IEEE}
}

@article{yamada2015static,
  title={Static grasp stability analysis of multiple spatial objects},
  author={Yamada, Takayoshi and Yamamoto, Hidehiko},
  journal={Journal of Control Science and Engineering},
  volume={3},
  pages={118--139},
  year={2015}
}

@inproceedings{harada1998kinematics,
  title={Kinematics and internal force in grasping multiple objects},
  author={Harada, Kensuke and Kaneko, Makoto},
  booktitle={Proceedings. 1998 IEEE/RSJ International Conference on Intelligent Robots and Systems. Innovations in Theory, Practice and Applications (Cat. No. 98CH36190)},
  volume={1},
  pages={298--303},
  year={1998},
  organization={IEEE}
}

@inproceedings{agboh2022multi,
  title={Multi-object grasping in the plane},
  author={Agboh, Wisdom C and Ichnowski, Jeffrey and Goldberg, Ken and Dogar, Mehmet R},
  booktitle={The International Symposium of Robotics Research},
  pages={222--238},
  year={2022},
  organization={Springer}
}

@inproceedings{sun2022multi,
  title={Multi-object grasping-types and taxonomy},
  author={Sun, Yu and Amatova, Eliza and Chen, Tianze},
  booktitle={2022 International conference on robotics and automation (ICRA)},
  pages={777--783},
  year={2022},
  organization={IEEE}
}

@article{yao2023exploiting,
  title={Exploiting kinematic redundancy for robotic grasping of multiple objects},
  author={Yao, Kunpeng and Billard, Aude},
  journal={IEEE Transactions on Robotics},
  volume={39},
  number={3},
  pages={1982--2002},
  year={2023},
  publisher={IEEE}
}

@article{liu2021synthesizing,
  title={Synthesizing diverse and physically stable grasps with arbitrary hand structures using differentiable force closure estimator},
  author={Liu, Tengyu and Liu, Zeyu and Jiao, Ziyuan and Zhu, Yixin and Zhu, Song-Chun},
  journal={IEEE Robotics and Automation Letters},
  volume={7},
  number={1},
  pages={470--477},
  year={2021},
  publisher={IEEE}
}

@article{ho2020denoising,
  title={Denoising diffusion probabilistic models},
  author={Ho, Jonathan and Jain, Ajay and Abbeel, Pieter},
  journal={Advances in neural information processing systems},
  volume={33},
  pages={6840--6851},
  year={2020}
}

@article{mnih2015human,
  title={Human-level control through deep reinforcement learning},
  author={Mnih, Volodymyr and Kavukcuoglu, Koray and Silver, David and Rusu, Andrei A and Veness, Joel and Bellemare, Marc G and Graves, Alex and Riedmiller, Martin and Fidjeland, Andreas K and Ostrovski, Georg and others},
  journal={nature},
  volume={518},
  number={7540},
  pages={529--533},
  year={2015},
  publisher={Nature Publishing Group}
}

@inproceedings{DBLP:journals/corr/LillicrapHPHETS15,
  author       = {Timothy P. Lillicrap and
                  Jonathan J. Hunt and
                  Alexander Pritzel and
                  Nicolas Heess and
                  Tom Erez and
                  Yuval Tassa and
                  David Silver and
                  Daan Wierstra},
  editor       = {Yoshua Bengio and
                  Yann LeCun},
  title        = {Continuous control with deep reinforcement learning},
  booktitle    = {4th International Conference on Learning Representations, {ICLR} 2016,
                  San Juan, Puerto Rico, May 2-4, 2016, Conference Track Proceedings},
  year         = {2016},
  url          = {http://arxiv.org/abs/1509.02971},
  bibsource    = {dblp computer science bibliography, https://dblp.org}
}

@article{schulman2017proximal,
  title={Proximal policy optimization algorithms},
  author={Schulman, John and Wolski, Filip and Dhariwal, Prafulla and Radford, Alec and Klimov, Oleg},
  journal={arXiv preprint arXiv:1707.06347},
  year={2017}
}

@inproceedings{haarnoja2018soft,
  title={Soft actor-critic: Off-policy maximum entropy deep reinforcement learning with a stochastic actor},
  author={Haarnoja, Tuomas and Zhou, Aurick and Abbeel, Pieter and Levine, Sergey},
  booktitle={International conference on machine learning},
  pages={1861--1870},
  year={2018},
  organization={Pmlr}
}

@inproceedings{xu2023unidexgrasp,
  title={Unidexgrasp: Universal robotic dexterous grasping via learning diverse proposal generation and goal-conditioned policy},
  author={Xu, Yinzhen and Wan, Weikang and Zhang, Jialiang and Liu, Haoran and Shan, Zikang and Shen, Hao and Wang, Ruicheng and Geng, Haoran and Weng, Yijia and Chen, Jiayi and others},
  booktitle={Proceedings of the IEEE/CVF Conference on Computer Vision and Pattern Recognition},
  pages={4737--4746},
  year={2023}
}

@inproceedings{wan2023unidexgrasp++,
  title={Unidexgrasp++: Improving dexterous grasping policy learning via geometry-aware curriculum and iterative generalist-specialist learning},
  author={Wan, Weikang and Geng, Haoran and Liu, Yun and Shan, Zikang and Yang, Yaodong and Yi, Li and Wang, He},
  booktitle={Proceedings of the IEEE/CVF International Conference on Computer Vision},
  pages={3891--3902},
  year={2023}
}

@inproceedings{huang2025efficient,
  title={Efficient residual learning with mixture-of-experts for universal dexterous grasping},
  author={Huang, Ziye and Yuan, Haoqi and Fu, Yuhui and Lu, Zongqing},
  booktitle={International Conference on Learning Representations},
  volume={2025},
  pages={87492--87510},
  year={2025}
}

@inproceedings{su2026rfs,
  title={Rfs: Reinforcement learning with residual flow steering for dexterous manipulation},
  author={Su, Entong and Westenbroek, Tyler and Nagabandi, Anusha and Gupta, Abhishek},
  booktitle={The Fourteenth International Conference on Learning Representations},
  year={2026}
}

@inproceedings{DBLP:conf/rss/HuMS24,
  author       = {Hengyuan Hu and
                  Suvir Mirchandani and
                  Dorsa Sadigh},
  editor       = {Dana Kulic and
                  Gentiane Venture and
                  Kostas E. Bekris and
                  Enrique Coronado},
  title        = {Imitation Bootstrapped Reinforcement Learning},
  booktitle    = {Robotics: Science and Systems XX, Delft, The Netherlands, July 15-19,
                  2024},
  year         = {2024},
  url          = {https://doi.org/10.15607/RSS.2024.XX.056},
  doi          = {10.15607/RSS.2024.XX.056},
  bibsource    = {dblp computer science bibliography, https://dblp.org}
}

@article{yuan2025demograsp,
  title={Demograsp: Universal dexterous grasping from a single demonstration},
  author={Yuan, Haoqi and Huang, Ziye and Wang, Ye and Mao, Chuan and Xu, Chaoyi and Lu, Zongqing},
  journal={arXiv preprint arXiv:2509.22149},
  year={2025}
}

@article{pomerleau1988alvinn,
  title={Alvinn: An autonomous land vehicle in a neural network},
  author={Pomerleau, Dean A},
  journal={Advances in neural information processing systems},
  volume={1},
  year={1988}
}

@inproceedings{ross2011reduction,
  title={A reduction of imitation learning and structured prediction to no-regret online learning},
  author={Ross, St{\'e}phane and Gordon, Geoffrey and Bagnell, Drew},
  booktitle={Proceedings of the fourteenth international conference on artificial intelligence and statistics},
  pages={627--635},
  year={2011},
  organization={JMLR Workshop and Conference Proceedings}
}

@inproceedings{ziebart2008maximum,
  title={Maximum entropy inverse reinforcement learning.},
  author={Ziebart, Brian D and Maas, Andrew L and Bagnell, J Andrew and Dey, Anind K and others},
  booktitle={Aaai},
  volume={8},
  pages={1433--1438},
  year={2008},
  organization={Chicago, IL, USA}
}

@inproceedings{ng2000algorithms,
  title={Algorithms for inverse reinforcement learning.},
  author={Ng, Andrew Y and Russell, Stuart and others},
  booktitle={Icml},
  volume={1},
  number={2},
  pages={2},
  year={2000}
}

@article{chi2025diffusion,
  title={Diffusion policy: Visuomotor policy learning via action diffusion},
  author={Chi, Cheng and Xu, Zhenjia and Feng, Siyuan and Cousineau, Eric and Du, Yilun and Burchfiel, Benjamin and Tedrake, Russ and Song, Shuran},
  journal={The International Journal of Robotics Research},
  volume={44},
  number={10-11},
  pages={1684--1704},
  year={2025},
  publisher={Sage Publications Sage UK: London, England}
}

@inproceedings{qin2022dexmv,
  title={Dexmv: Imitation learning for dexterous manipulation from human videos},
  author={Qin, Yuzhe and Wu, Yueh-Hua and Liu, Shaowei and Jiang, Hanwen and Yang, Ruihan and Fu, Yang and Wang, Xiaolong},
  booktitle={European Conference on Computer Vision},
  pages={570--587},
  year={2022},
  organization={Springer}
}

@article{qin2023anyteleop,
  title={Anyteleop: A general vision-based dexterous robot arm-hand teleoperation system},
  author={Qin, Yuzhe and Yang, Wei and Huang, Binghao and Van Wyk, Karl and Su, Hao and Wang, Xiaolong and Chao, Yu-Wei and Fox, Dieter},
  journal={arXiv preprint arXiv:2307.04577},
  year={2023}
}

@article{calli2017yale,
  title={Yale-CMU-Berkeley dataset for robotic manipulation research},
  author={Calli, Berk and Singh, Arjun and Bruce, James and Walsman, Aaron and Konolige, Kurt and Srinivasa, Siddhartha and Abbeel, Pieter and Dollar, Aaron M},
  journal={The International Journal of Robotics Research},
  volume={36},
  number={3},
  pages={261--268},
  year={2017},
  publisher={SAGE Publications Sage UK: London, England}
}

@inproceedings{brahmbhatt2019contactdb,
  title={Contactdb: Analyzing and predicting grasp contact via thermal imaging},
  author={Brahmbhatt, Samarth and Ham, Cusuh and Kemp, Charles C and Hays, James},
  booktitle={Proceedings of the IEEE/CVF conference on computer vision and pattern recognition},
  pages={8709--8719},
  year={2019}
}

@article{makoviychuk2021isaac,
  title={Isaac gym: High performance gpu-based physics simulation for robot learning},
  author={Makoviychuk, Viktor and Wawrzyniak, Lukasz and Guo, Yunrong and Lu, Michelle and Storey, Kier and Macklin, Miles and Hoeller, David and Rudin, Nikita and Allshire, Arthur and Handa, Ankur and others},
  journal={arXiv preprint arXiv:2108.10470},
  year={2021}
}

@inproceedings{qi2017pointnet,
  title={Pointnet: Deep learning on point sets for 3d classification and segmentation},
  author={Qi, Charles R and Su, Hao and Mo, Kaichun and Guibas, Leonidas J},
  booktitle={Proceedings of the IEEE conference on computer vision and pattern recognition},
  pages={652--660},
  year={2017}
}

@article{clevert2015fast,
  title={Fast and accurate deep network learning by exponential linear units (elus)},
  author={Clevert, Djork-Arn{\'e} and Unterthiner, Thomas and Hochreiter, Sepp},
  journal={arXiv preprint arXiv:1511.07289},
  volume={4},
  number={5},
  pages={11},
  year={2015}
}

@article{wang2022dexgraspnet,
  title={Dexgraspnet: A large-scale robotic dexterous grasp dataset for general objects based on simulation},
  author={Wang, Ruicheng and Zhang, Jialiang and Chen, Jiayi and Xu, Yinzhen and Li, Puhao and Liu, Tengyu and Wang, He},
  journal={arXiv preprint arXiv:2210.02697},
  year={2022}
}

@article{bjorck2025gr00t,
  title={Gr00t n1: An open foundation model for generalist humanoid robots},
  author={Bjorck, Johan and Casta{\~n}eda, Fernando and Cherniadev, Nikita and Da, Xingye and Ding, Runyu and Fan, Linxi and Fang, Yu and Fox, Dieter and Hu, Fengyuan and Huang, Spencer and others},
  journal={arXiv preprint arXiv:2503.14734},
  year={2025}
}

@article{lipman2022flow,
  title={Flow matching for generative modeling},
  author={Lipman, Yaron and Chen, Ricky TQ and Ben-Hamu, Heli and Nickel, Maximilian and Le, Matt},
  journal={arXiv preprint arXiv:2210.02747},
  year={2022}
}

@article{xu2025dexumi,
  title={Dexumi: Using human hand as the universal manipulation interface for dexterous manipulation},
  author={Xu, Mengda and Zhang, Han and Hou, Yifan and Xu, Zhenjia and Fan, Linxi and Veloso, Manuela and Song, Shuran},
  journal={arXiv preprint arXiv:2505.21864},
  year={2025}
}

@inproceedings{liang2025dexhanddiff,
  title={Dexhanddiff: Interaction-aware diffusion planning for adaptive dexterous manipulation},
  author={Liang, Zhixuan and Mu, Yao and Wang, Yixiao and Chen, Tianxing and Shao, Wenqi and Zhan, Wei and Tomizuka, Masayoshi and Luo, Ping and Ding, Mingyu},
  booktitle={Proceedings of the Computer Vision and Pattern Recognition Conference},
  pages={1745--1755},
  year={2025}
}

@inproceedings{DBLP:conf/rss/SiZTK24,
  author       = {Zilin Si and
                  Kevin Lee Zhang and
                  Zeynep Temel and
                  Oliver Kroemer},
  editor       = {Dana Kulic and
                  Gentiane Venture and
                  Kostas E. Bekris and
                  Enrique Coronado},
  title        = {Tilde: Teleoperation for Dexterous In-Hand Manipulation Learning with
                  a DeltaHand},
  booktitle    = {Robotics: Science and Systems XX, Delft, The Netherlands, July 15-19,
                  2024},
  year         = {2024},
  url          = {https://doi.org/10.15607/RSS.2024.XX.128},
  doi          = {10.15607/RSS.2024.XX.128},
  bibsource    = {dblp computer science bibliography, https://dblp.org}
}

@inproceedings{DBLP:conf/rss/ZeZZHWX24,
  author       = {Yanjie Ze and
                  Gu Zhang and
                  Kangning Zhang and
                  Chenyuan Hu and
                  Muhan Wang and
                  Huazhe Xu},
  editor       = {Dana Kulic and
                  Gentiane Venture and
                  Kostas E. Bekris and
                  Enrique Coronado},
  title        = {3D Diffusion Policy: Generalizable Visuomotor Policy Learning via
                  Simple 3D Representations},
  booktitle    = {Robotics: Science and Systems XX, Delft, The Netherlands, July 15-19,
                  2024},
  year         = {2024},
  url          = {https://doi.org/10.15607/RSS.2024.XX.067},
  doi          = {10.15607/RSS.2024.XX.067},
  bibsource    = {dblp computer science bibliography, https://dblp.org}
}

@article{argall2009survey,
  title={A survey of robot learning from demonstration},
  author={Argall, Brenna D and Chernova, Sonia and Veloso, Manuela and Browning, Brett},
  journal={Robotics and autonomous systems},
  volume={57},
  number={5},
  pages={469--483},
  year={2009},
  publisher={Elsevier}
}

@article{osa2018algorithmic,
  title={An algorithmic perspective on imitation learning},
  author={Osa, Takayuki and Pajarinen, Joni and Neumann, Gerhard and Bagnell, J Andrew and Abbeel, Pieter and Peters, Jan},
  journal={Foundations and Trends{\textregistered} in Robotics},
  volume={7},
  number={1-2},
  pages={1--179},
  year={2018},
  publisher={Emerald Publishing Limited}
}

@article{hussein2017imitation,
  title={Imitation learning: A survey of learning methods},
  author={Hussein, Ahmed and Gaber, Mohamed Medhat and Elyan, Eyad and Jayne, Chrisina},
  journal={ACM Computing Surveys (CSUR)},
  volume={50},
  number={2},
  pages={1--35},
  year={2017},
  publisher={ACM New York, NY, USA}
}

@article{bishop1994mixture,
  title={Mixture density networks},
  author={Bishop, Christopher M},
  year={1994},
  publisher={Aston University}
}

@inproceedings{DBLP:conf/cvpr/ZhongJYM25,
  author       = {Yiming Zhong and
                  Qi Jiang and
                  Jingyi Yu and
                  Yuexin Ma},
  title        = {DexGrasp Anything: Towards Universal Robotic Dexterous Grasping with
                  Physics Awareness},
  booktitle    = {{IEEE/CVF} Conference on Computer Vision and Pattern Recognition,
                  {CVPR} 2025, Nashville, TN, USA, June 11-15, 2025},
  pages        = {22584--22594},
  publisher    = {Computer Vision Foundation / {IEEE}},
  year         = {2025},
  url          = {https://openaccess.thecvf.com/content/CVPR2025/html/Zhong\_DexGrasp\_Anything\_Towards\_Universal\_Robotic\_Dexterous\_Grasping\_with\_Physics\_Awareness\_CVPR\_2025\_paper.html},
  doi          = {10.1109/CVPR52734.2025.02103},
  bibsource    = {dblp computer science bibliography, https://dblp.org}
}

@inproceedings{DBLP:conf/eccv/ZhangHPWHCZD24,
  author       = {Jiyao Zhang and
                  Weiyao Huang and
                  Bo Peng and
                  Mingdong Wu and
                  Fei Hu and
                  Zijian Chen and
                  Bo Zhao and
                  Hao Dong},
  editor       = {Ales Leonardis and
                  Elisa Ricci and
                  Stefan Roth and
                  Olga Russakovsky and
                  Torsten Sattler and
                  G{\"{u}}l Varol},
  title        = {Omni6DPose: {A} Benchmark and Model for Universal 6D Object Pose Estimation
                  and Tracking},
  booktitle    = {Computer Vision - {ECCV} 2024 - 18th European Conference, Milan, Italy,
                  September 29-October 4, 2024, Proceedings, Part {LXXV}},
  series       = {Lecture Notes in Computer Science},
  pages        = {199--216},
  publisher    = {Springer},
  year         = {2024},
  url          = {https://doi.org/10.1007/978-3-031-73226-3\_12},
  doi          = {10.1007/978-3-031-73226-3\_12},
  bibsource    = {dblp computer science bibliography, https://dblp.org}
}

@article{DBLP:journals/scirobotics/ChenTWKAA23,
  author       = {Tao Chen and
                  Megha Tippur and
                  Siyang Wu and
                  Vikash Kumar and
                  Edward H. Adelson and
                  Pulkit Agrawal},
  title        = {Visual dexterity: In-hand reorientation of novel and complex object
                  shapes},
  journal      = {Sci. Robotics},
  volume       = {8},
  number       = {84},
  year         = {2023},
  url          = {https://doi.org/10.1126/scirobotics.adc9244},
  doi          = {10.1126/SCIROBOTICS.ADC9244},
  bibsource    = {dblp computer science bibliography, https://dblp.org}
}

@article{DBLP:journals/ral/MorrisonCL20,
  author       = {Douglas Morrison and
                  Peter Corke and
                  J{\"{u}}rgen Leitner},
  title        = {EGAD! An Evolved Grasping Analysis Dataset for Diversity and Reproducibility
                  in Robotic Manipulation},
  journal      = {{IEEE} Robotics Autom. Lett.},
  volume       = {5},
  number       = {3},
  pages        = {4368--4375},
  year         = {2020},
  url          = {https://doi.org/10.1109/LRA.2020.2992195},
  doi          = {10.1109/LRA.2020.2992195},
  bibsource    = {dblp computer science bibliography, https://dblp.org}
}

@article{he2025sequential,
  title={Sequential Multi-Object Grasping with One Dexterous Hand},
  author={He, Sicheng and Shangguan, Zeyu and Wang, Kuanning and Gu, Yongchong and Fu, Yuqian and Fu, Yanwei and Seita, Daniel},
  journal={arXiv preprint arXiv:2503.09078},
  year={2025}
}

\clearpage
\appendix
\raggedbottom
\setlength{\intextsep}{8pt plus 2pt minus 2pt}
\setlength{\textfloatsep}{10pt plus 2pt minus 2pt}

\section{Implementation Details}
\subsection{Datasets and Task Setup}


\begin{figure}[H]
    \centering
    \includegraphics[width=\linewidth, trim=7pt 9pt 7pt 10pt, clip]{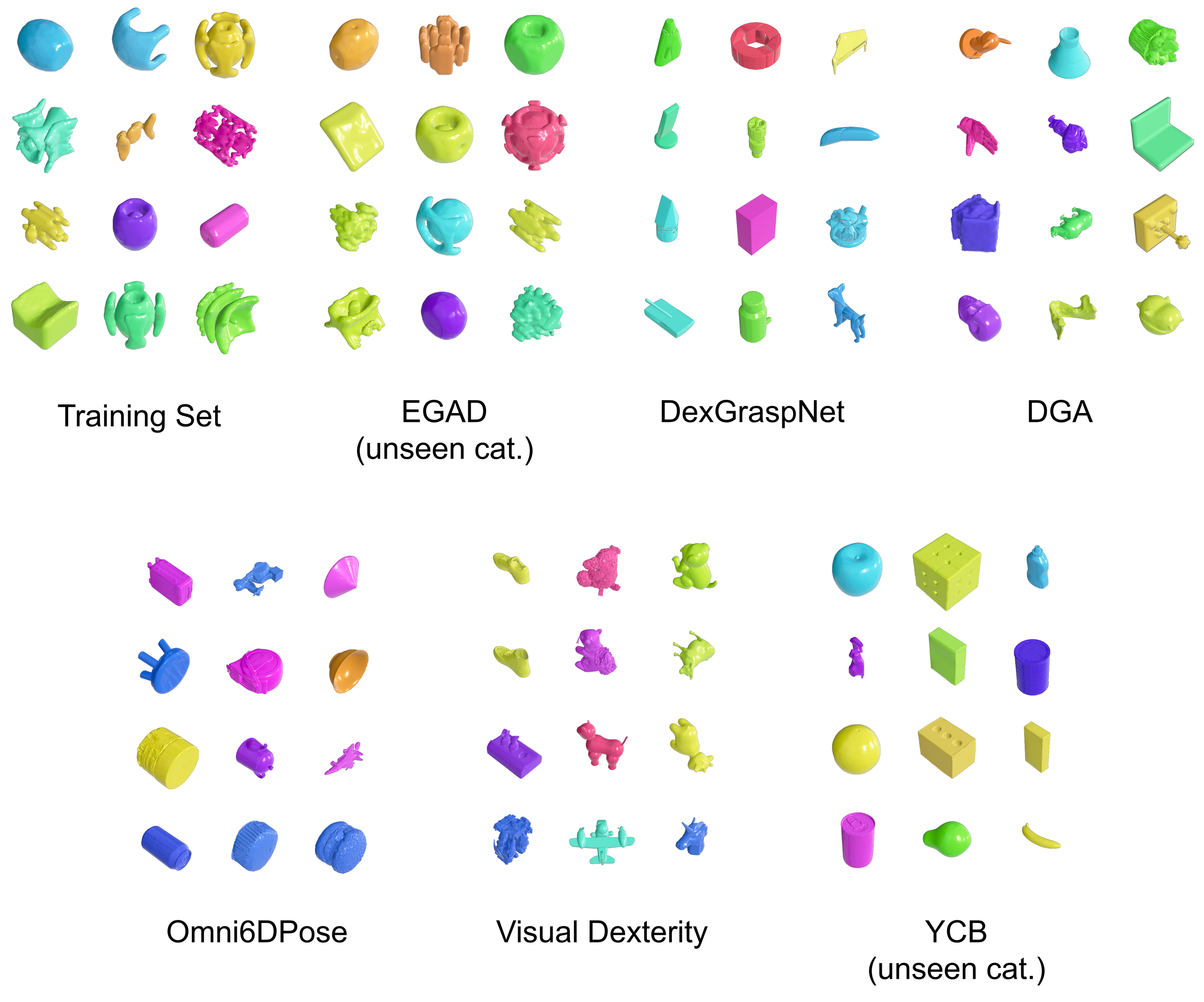}
    \caption{Representative objects randomly sampled from the training set and the six evaluation datasets used in our benchmark for visualization.}
    \label{fig:datasets}
\end{figure}

\subsubsection{Dataset Composition}
To align with the experimental setup of MultiGrasp~\cite{li2024grasp}, unless specified, our training set comprises 8 distinct objects, with an additional 3 objects reserved for out-of-distribution (OOD) generalization testing. In total, the training configurations encompass 28 unique object pairs with 35 stable orientations. To rigorously evaluate generalization capabilities, we hold out 9 stable orientations for \textit{Unseen Place.} testing and 8 object pairs for \textit{Unseen Comb.} testing. Furthermore, the complete evaluation suite introduces 3 novel objects, forming 6 distinct unseen object combinations with a total of 71 stable orientations.

We additionally construct an extended object set containing 18 objects. This set consists of the original 11 objects used in MultiGrasp~\cite{li2024grasp}, and 7 objects selected from YCB~\cite{calli2017yale}. The resulting object pool provides substantially greater diversity in size, geometry and aspect ratio than the original training set. Meanwhile, to further scale up the diversity and complexity of the training distribution, we expand our training set using the EGAD dataset~\cite{DBLP:journals/ral/MorrisonCL20}. Specifically, we select 3 distinct objects from each of the 24 difficulty-based categories defined in the original dataset, resulting in a pool of 72 unique training objects. This selection yields exactly 2,628 unique pairwise combinations utilized during the training phase. For each object, we pre-collect a dedicated set of 100 stable orientations.

To rigorously challenge the model's out-of-distribution generalization ability, we construct an extensive evaluation suite spanning six diverse object datasets (Figure~\ref{fig:datasets}): Visual Dexterity~\cite{DBLP:journals/scirobotics/ChenTWKAA23}, EGAD~\cite{DBLP:journals/ral/MorrisonCL20}, YCB~\cite{calli2017yale}, DexGraspNet~\citep{wang2022dexgraspnet}, Omni6DPose~\cite{DBLP:conf/eccv/ZhangHPWHCZD24}, and DexGrasp Anything (DGA)~\cite{DBLP:conf/cvpr/ZhongJYM25}. We filter out geometric outliers by enforcing an aspect ratio under 3 and uniformly rescale the maximum dimension of each object to $4.5\,\text{cm}$. Evaluation is conducted exclusively using a separate, disjoint set of 100 pre-collected stable orientations per object that is completely isolated from the training distribution. Our testing protocol fixes the random seed at 42 across 5 independent rounds, ensuring each unique object combination is sampled with equal frequency. To maintain a balanced evaluation scale, we employ stratified sampling to select exactly 48 objects for each dataset, with the exception of YCB, which consists of 12 manually curated exemplars tailored for dexterous manipulation. Critically, to evaluate intra-dataset instance generalization, the 48 testing objects from EGAD are strictly disjoint from the training partition, obtained by sampling 2 novel objects from each of the 24 native complexity levels. For the remaining datasets: the 48 objects for Visual Dexterity are sampled evenly across three aspect ratio intervals ($1.0\text{--}1.5$, $1.5\text{--}2.0$, and $2.0\text{--}3.0$); For DexGraspNet and Omni6DPose, objects are selected through a semantic retrieval process that matches object categories to those represented in the training set, ensuring comparable category coverage; and DGA objects are drawn directly from its predefined testing split.

\subsubsection{Pose Initialization} 
Two objects are placed within the spatial boundaries defined by their corresponding layout configurations. Specifically, for the proximal layout, objects are randomly distributed within a $7.5 \times 7.5\,\text{cm}^2$ square bounding region. On the other hand, the separated layout enforces an explicit spatial separation between the two objects, with their center-to-center distance predefined at $15\,\text{cm}$, $20\,\text{cm}$, or $25\,\text{cm}$. To ensure physical stability, each object is initialized with an orientation randomly sampled from a predefined set of stable poses. To avoid inter-object overlap, we introduce a point-cloud-based penetration metric. Specifically, the point clouds of both objects are transformed into the world frame, and a penetration penalty is computed as the average inward distance of points from one object that lie inside the axis-aligned bounding box (AABB) of the other object. A layout configuration is accepted only if this cumulative penalty is below a threshold. If an invalid overlap is detected, the spatial poses of both objects are jointly re-sampled iteratively until a valid, collision-free configuration is achieved.
\subsection{RL Policy Training}
\label{appendix:rl}              
\subsubsection{Action} 
The raw residual $\mathbf{r}=(\Delta\mathbf{x}, \Delta\boldsymbol{\phi}, \Delta\boldsymbol{\theta}) \in [-1, 1]$ is rescaled component-wise to the following physical ranges: wrist translation $\Delta\mathbf{x'} \in [-0.05, 0.05]^3$\,m, wrist Euler-angle offset $\Delta\boldsymbol{\phi'} \in [-1.57, 1.57]^3$\,rad, and hand joint offset $\Delta\boldsymbol{\theta'} \in [-1, 1]$\,rad. For the multi-stage strategy, the residual is split into two stage-specific sets acting on disjoint finger groups. The first-stage residual $(\Delta\mathbf{x'_1}, \Delta\boldsymbol{\phi'_1}, \Delta\boldsymbol{\theta'_1})$ controls the wrist and the first finger group (thumb and index) to grasp $O_1$, while the remaining fingers stay at their initial pose. The second-stage residual $(\Delta\mathbf{x'_2}, \Delta\boldsymbol{\phi'_2}, \Delta\boldsymbol{\theta'_2})$ controls the wrist and the second finger group (the remaining fingers) to grasp $O_2$, while the first group is locked. Here $\Delta\boldsymbol{\theta'_1}$ and $\Delta\boldsymbol{\theta'_2}$ contain the joint dimensions of their respective finger groups and together span all actuated hand joints; the exact partition is embodiment-dependent. For a hand with $d_h$ actuated joints, the total residual dimension is therefore $12 + d_h$ (two $6$-D wrist residuals plus the full set of hand joints).

\subsubsection{Network Architecture} 
The actor and critic share the same backbone: each per-object point cloud, down-sampled to $256$ points, is encoded by a simplified PointNet~\citep{qi2017pointnet} into a $128$-dimensional feature; these features are concatenated with the other features and fed to two separate MLP heads of widths $[1024, 1024, 512, 512]$ with ELU activations~\citep{clevert2015fast}. 

\subsubsection{Training Hyperparameters} 
All state-based policies are trained with PPO~\citep{schulman2017proximal} under the single-step formulation (one rollout step per iteration), using the Adam optimizer and an adaptive learning-rate schedule driven by a target KL divergence. Training is run in IsaacGym with $8{,}000$ parallel environments on a single GPU. The training hyperparameters are listed in Table~\ref{tab:ppo_hparams}. 

\begin{table}[htpb]
\centering
\caption{RL hyperparameters used to train all state-based policies.}
\label{tab:ppo_hparams}
\small
\begin{tabular}{ll}
\toprule
Hyperparameter & Value \\
\midrule
Optimizer                       & Adam \\
Initial learning rate           & $3\!\times\!10^{-4}$ \\
Target KL                       & $0.016$ \\
Clip range                      & $0.2$ \\
Entropy coefficient             & $0$ \\
Value-loss coefficient          & $2.0$ \\
Max gradient norm               & $1.0$ \\
Optimization epochs per iteration      & $5$ \\
Minibatches per epoch              & $4$ \\
Rollout steps per iteration            & $1$ \\
Action clip range               & $[-1,1]$ \\
Observation clip range          & $[-5,5]$ \\
\bottomrule
\end{tabular}
\end{table}                                           

\subsubsection{Simulation and Control} 
We use IsaacGym~\citep{makoviychuk2021isaac} with a physics timestep of $\Delta t_\text{sim}{=}1/60$\,s and the PhysX TGS solver ($2$ substeps, $8$ position iterations, $0$ velocity iterations). Each control step advances $20$ physics steps; within a control step the controller linearly interpolates the joint position targets between the previous and current setpoints across the $20$ physics steps. Both the arm and the hand are position-controlled. The per-step angular velocity is capped at $1.57$\,rad/s for the arm and $6.28$\,rad/s for the hand. The episode length is $40$ control steps for the single-shot strategy and $80$ control steps for the multi-stage strategy, accommodating the additional reach--grasp--lift cycle of the second object.

\subsubsection{Domain Randomization} 
At every episode reset we randomize each object's pose to promote robustness across placements. Its orientation is set to a stable resting rotation sampled uniformly from a precomputed per-object pool and further composed with a uniform random yaw about the vertical axis, while its planar position is perturbed by a uniform offset within a $0.075\,\text{m}$ square centered at its nominal location. To realize the separated regime defined in Sec.~\ref{subsec:problem} ($d_\text{min}\geq d_\text{thresh}$), we enforce a minimum separation: whenever the sampled $d_\text{min}$ falls below $d_\text{thresh}$, the two objects are pushed symmetrically apart along their connecting line until $d_\text{min}{=}d_\text{thresh}$.

\subsection{Vision-based Policy Training}
\label{appendix:vla}

\subsubsection{Data Collection} 
We roll out the trained state-based policy in IsaacGym with visual randomization and retain only the successful episodes. The resulting dataset contains $30{,}000$ episodes ($1.2$M frames in total) recorded at $3$\,Hz. Each frame consists of two $256{\times}256$ RGB images from two cameras, a $13$-dimensional proprioceptive state, and the corresponding $13$-dimensional low-level command. The proprioceptive state comprises the $7$D arm end-effector pose and the $6$ active joint angles of the Inspire hand, whose remaining six finger joints are passively coupled to these actuated joints through fixed ratios; the command consists of the target end-effector pose and the target active hand joint angles. Beyond the object pose randomization used during state-based training (Appendix~\ref{appendix:rl}), we additionally apply visual randomization at every reset: (i) camera position and orientation perturbations of $\pm 0.02$\,m and $\pm 0.02$\,rad per axis; (ii) table-position shifts of $\pm 0.05$\,m in $x$ and $\pm 0.02$\,m in $y$; (iii) three directional lights with randomized directions, each with a directional intensity and an ambient term sampled uniformly from $[0.1, 0.8]$; and (iv) per-object appearance randomization, either a color drawn from a $21$-color palette or a random texture.

\begin{figure}[H]
    \centering
    \includegraphics[width=0.72\linewidth]{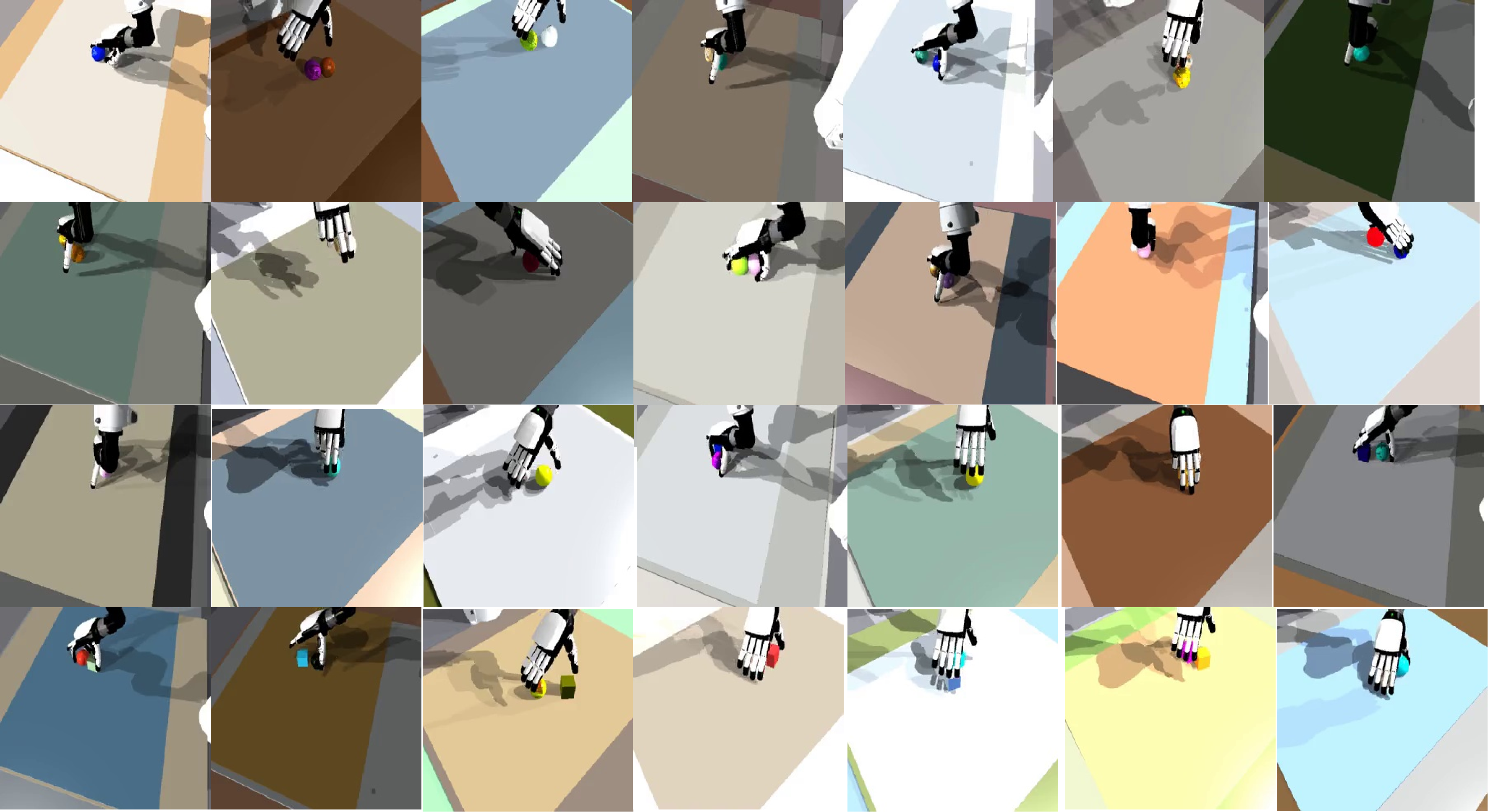}
    \caption{Examples of RGB observations used for vision-based policy training. The collected episodes are rendered with visual randomization, including perturbations of camera pose, table position, lighting conditions, object colors, and textures, so that the policy is exposed to diverse viewpoints and appearance variations during training.}
    \label{fig:sim_data}
\end{figure}

\subsubsection{Model Architecture} 
We use GR00T-N1.5-3B~\citep{bjorck2025gr00t} as the vision-language-action policy. At each control step, the model is conditioned on three inputs: two RGB observations from the dual-camera setup, resized from $256{\times}256$ to $224{\times}224$; the $13$-dimensional proprioceptive state; and a natural-language task description. It predicts an action chunk of horizon $H{=}10$, where each action is a $13$-dimensional low-level command composed of a $7$D target end-effector pose and $6$ target active hand joint angles. The state and action vectors are min--max normalized and zero-padded to match the model's maximum input widths, respectively. During deployment, we execute the policy in a receding-horizon fashion: after each inference call, only the first $5$ actions of the predicted chunk are applied before the model replans from the latest observations.

\subsubsection{Training Details} 
We fully fine-tune the model on $8$ NVIDIA RTX 4090 GPUs. All training hyperparameters are listed in Table~\ref{tab:vla_hparams}.

\begin{table}[h]
\centering
\caption{Hyperparameters for fine-tuning the vision-based policy.}
\label{tab:vla_hparams}
\small
\begin{tabular}{ll}
\toprule
Hyperparameter & Value \\
\midrule
Training steps                  & $100{,}000$ \\
Per-GPU batch size              & $4$ \\
Gradient accumulation steps          & $4$ \\
Effective batch size            & $128$ \\
Optimizer                       & AdamW \\
Peak learning rate              & $1\times10^{-4}$ \\
Learning-rate schedule          & cosine decay \\
Warmup ratio                    & $0.05$ \\
Weight decay                    & $1\times10^{-5}$ \\
Mixed precision                 & bfloat16 \\
Observation history             & $1$ \\
Action horizon                  & $10$ \\
Inference denoising steps       & $4$ \\
\bottomrule
\end{tabular}
\end{table} 

\section{Additional Simulation Results}
\subsection{Quantitative Results}

\begin{table}[H]
\centering
\caption{Success rates (\%) of diverse robot embodiments evaluated on In-Distribution (ID) and Out-of-Distribution (OOD) settings.}
\label{tab:grasp_results}
\renewcommand{\arraystretch}{1.35} 
\resizebox{\textwidth}{!}{%
\begin{tabular}{l c c c c c c c}
\toprule
& \textbf{In-Distribution} & \multicolumn{6}{c}{\textbf{Out-of-Distribution Suite}} \\
\cmidrule(lr){2-2} \cmidrule(lr){3-8}
\textbf{Embodiment} & Training Set & \begin{tabular}[c]{@{}c@{}}Visual\\ Dexterity\end{tabular} & DGA & YCB & EGAD & \begin{tabular}[c]{@{}c@{}}Omni6DPose\end{tabular} & \begin{tabular}[c]{@{}c@{}}DexGraspNet\end{tabular} \\
\midrule
Shadow                  & 54.90\% & \textbf{83.35\%} & \textbf{58.04\%} & \textbf{55.67\%} & 52.71\%          & \textbf{71.21\%} & \textbf{56.26\%} \\
WUJI                         & 40.13\% & 38.00\%          & 24.46\%          & 34.85\%          & 38.12\%          & 36.28\%          & 22.39\%          \\
FR3-Shadow                   & 50.90\% & 81.06\%          & 49.72\%          & 47.57\%          & 48.41\%          & 67.83\%          & 53.99\%          \\
UR5-Allegro                  & \textbf{59.48\%} & 63.09\%          & 32.94\%          & 52.81\%          & \textbf{57.24\%} & 59.25\%          & 33.09\%          \\
\bottomrule
\end{tabular}
}
\end{table}

Table~\ref{tab:grasp_results} reports the quantitative evaluation of the policies trained on 72 objects across diverse robot embodiments and six unseen object datasets. The empirical results reveal three critical insights regarding geometric generalization and morphological inductive biases:

\textbf{Morphological Advantage of Anthropomorphic Hands.}
Despite its moderate in-distribution (ID) training success rate ($54.90\%$), the standalone Shadow Hand configuration exhibits superior generalization, outperforming the other embodiments across five of the six out-of-distribution (OOD) testing suites. In particular, it achieves peak success rates of $83.35\%$ on Visual Dexterity and $71.21\%$ on Omni6DPose. These results highlight the morphological advantage of anthropomorphic hands: the highly articulated five-fingered structure provides a strong geometric inductive bias that allows the policy to adapt more effectively to severe distribution shifts and previously unseen object geometries.

\textbf{In-Distribution Calibration vs. OOD Robustness.}
Crucially, our results show that superior in-distribution (ID) performance does not necessarily translate into stronger cross-dataset generalization. The UR5-Allegro configuration achieves the highest training success rate ($59.48\%$) and generalizes well to the geometrically diverse EGAD dataset ($57.24\%$). However, its performance drops sharply on other benchmarks, failing to exceed a $34\%$ success rate on both DGA and DexGraspNet. This discrepancy suggests that the policy optimized for the four-fingered Allegro hand may overfit to specific interaction patterns in the training distribution, thereby exposing its vulnerability to unseen object geometries and semantic variations.

\textbf{Kinematic Trade-offs in Arm-Hand Systems.}
Comparing the standalone Shadow Hand with the integrated FR3-Shadow system reveals the performance cost introduced by full-arm deployment. Across all six evaluation benchmarks, incorporating the FR3 manipulator leads to consistent performance degradation, as illustrated by the drop from $58.04\%$ to $49.72\%$ on DGA. This systematic decline can be primarily attributed to the enlarged joint space and the coupled kinematic constraints introduced by the arm-hand system, which increase the complexity of policy search and limit the effective reachability and contact accessibility of the dexterous fingertips.

\subsection{Qualitative Results}

\subsubsection{Proximal Layouts.}

\begin{figure}[H]
\centering
\includegraphics[width=0.55\linewidth, trim=7pt 9pt 7pt 10pt, clip]{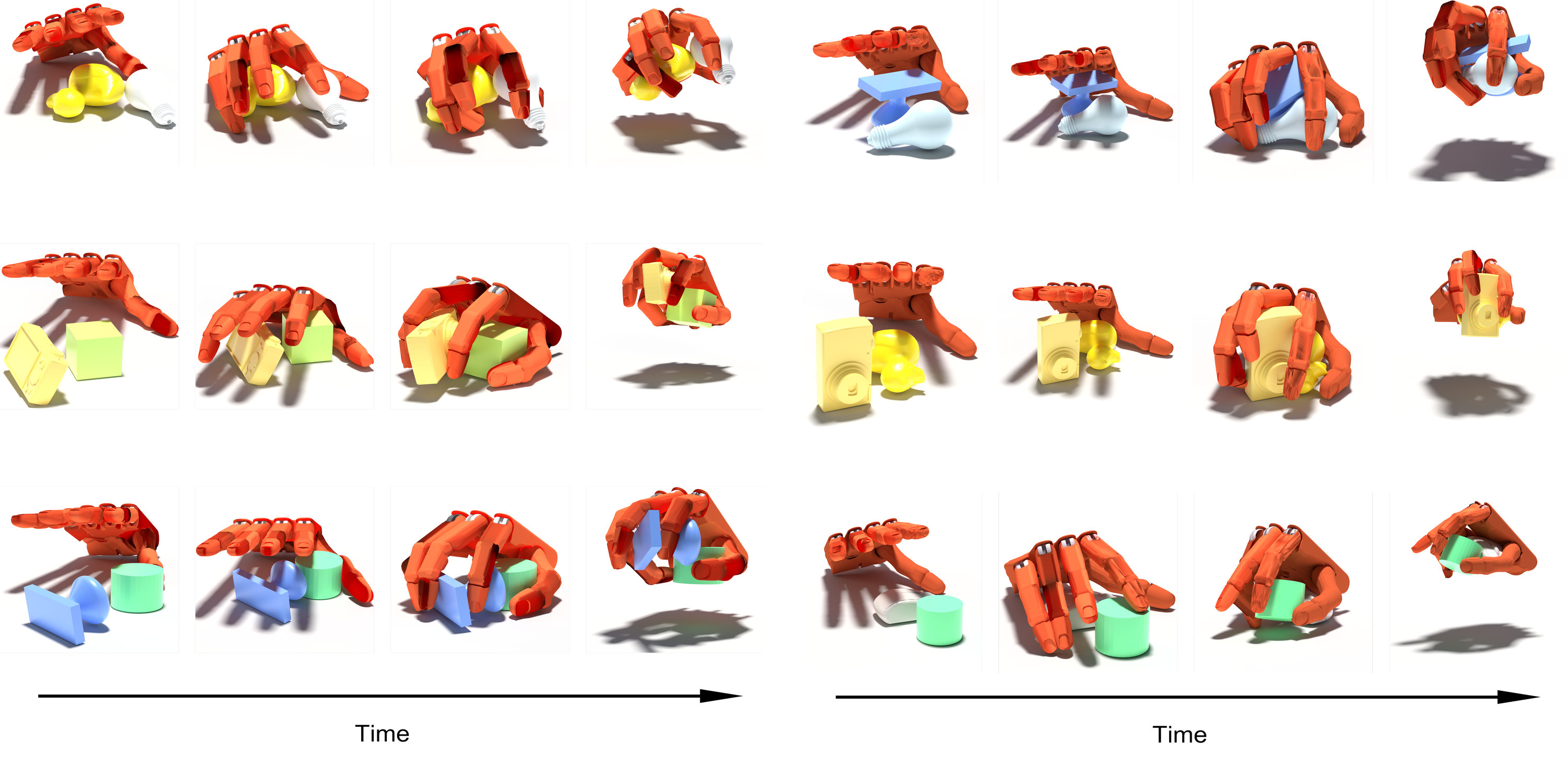}
\caption{\textbf{Proximal layouts.} Representative grasping trajectories under proximal object layouts in simulation. The visualizations are rendered using Blender.}
\label{fig:proximal_grasping}
\end{figure}
Figure~\ref{fig:proximal_grasping} shows representative grasping trajectories under proximal object layouts. When two objects are placed in close spatial proximity, the policy consistently develops a joint enclosure strategy that leverages the geometric coupling between the objects. Rather than treating the two targets as independent grasping instances, the fingers adaptively arrange around the combined object configuration and establish coordinated contacts across both objects before lifting. Notably, the learned behavior does not depend on a fixed grasp pattern. Across different object geometries and placements, the policy adjusts finger allocation and contact locations accordingly. This adaptive behavior suggests that the policy captures high-level geometric features of the objects, instead of merely memorizing object-specific grasp configurations.

\subsubsection{Separated Layouts.}

\begin{figure}[H]
\centering
\includegraphics[width=0.55\linewidth, trim=7pt 9pt 7pt 10pt, clip]{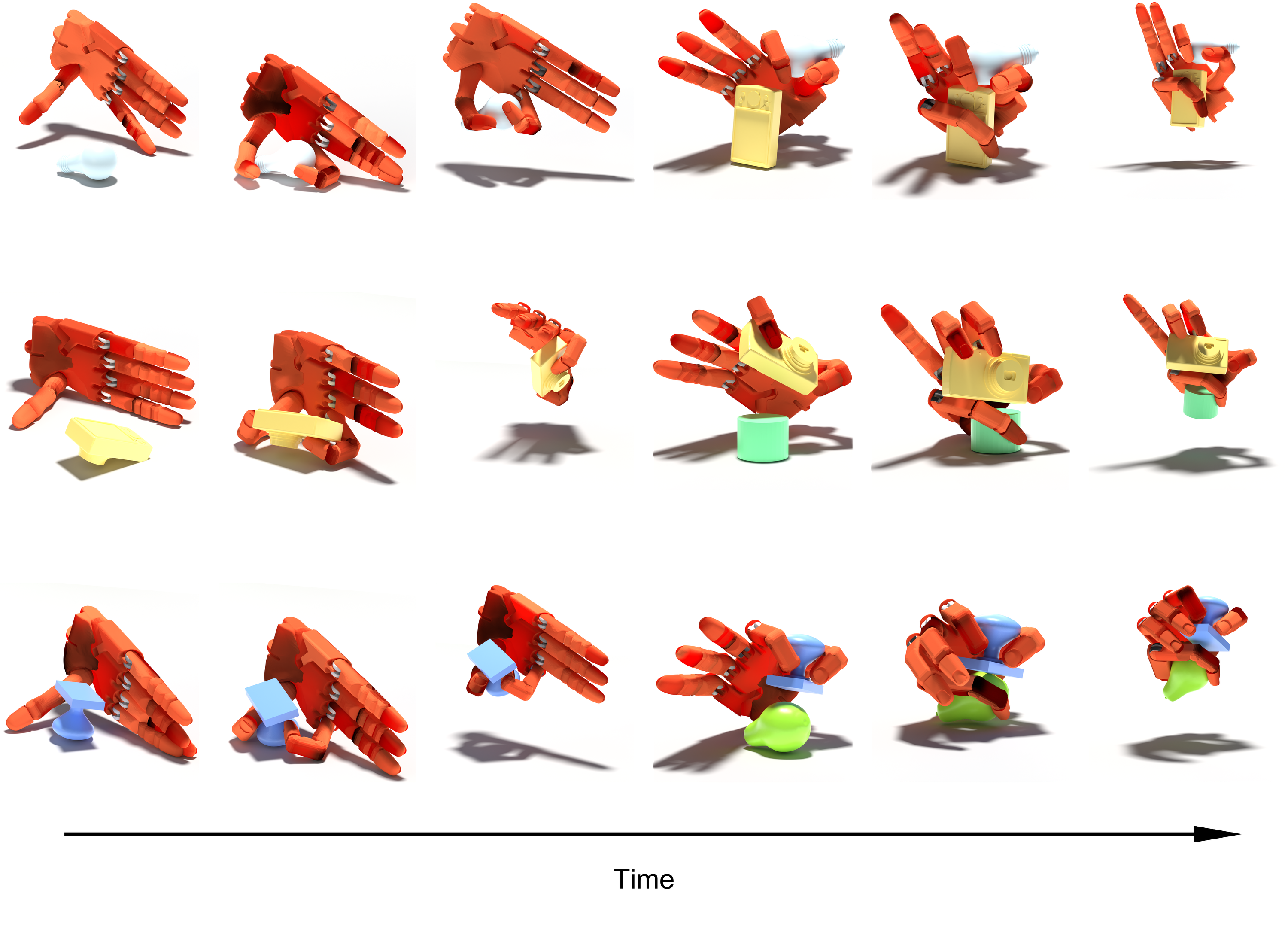}
\caption{\textbf{Separated layouts.} Representative grasping trajectories under separated object layouts in simulation. The visualizations are rendered using Blender.}
\label{fig:separated_grasping}
\end{figure}

Figure~\ref{fig:separated_grasping} illustrates representative grasping trajectories under separated layouts, where the two target objects are positioned beyond the range of a single enclosure. In such cases, the policy switches to a staged manipulation strategy. It first establishes a stable partial grasp on one object using a subset of fingers, while preserving sufficient hand mobility for subsequent interaction. The remaining fingers are then reconfigured to approach and acquire the second object, after which the hand executes a coordinated lifting motion to grasp both objects simultaneously. MoDex decomposes the dual-object grasping task into a sequence of coordinated sub-goals and adapts the finger allocation. These qualitative results further demonstrate the effectiveness of the proposed spatially adaptive strategy in handling object configurations that are infeasible for conventional single-shot grasping approaches.

\clearpage
\section{Real-World Experiments}
\subsection{Hardware Setup}

\begin{figure}[H]
\centering
    \includegraphics[width=0.55\linewidth]{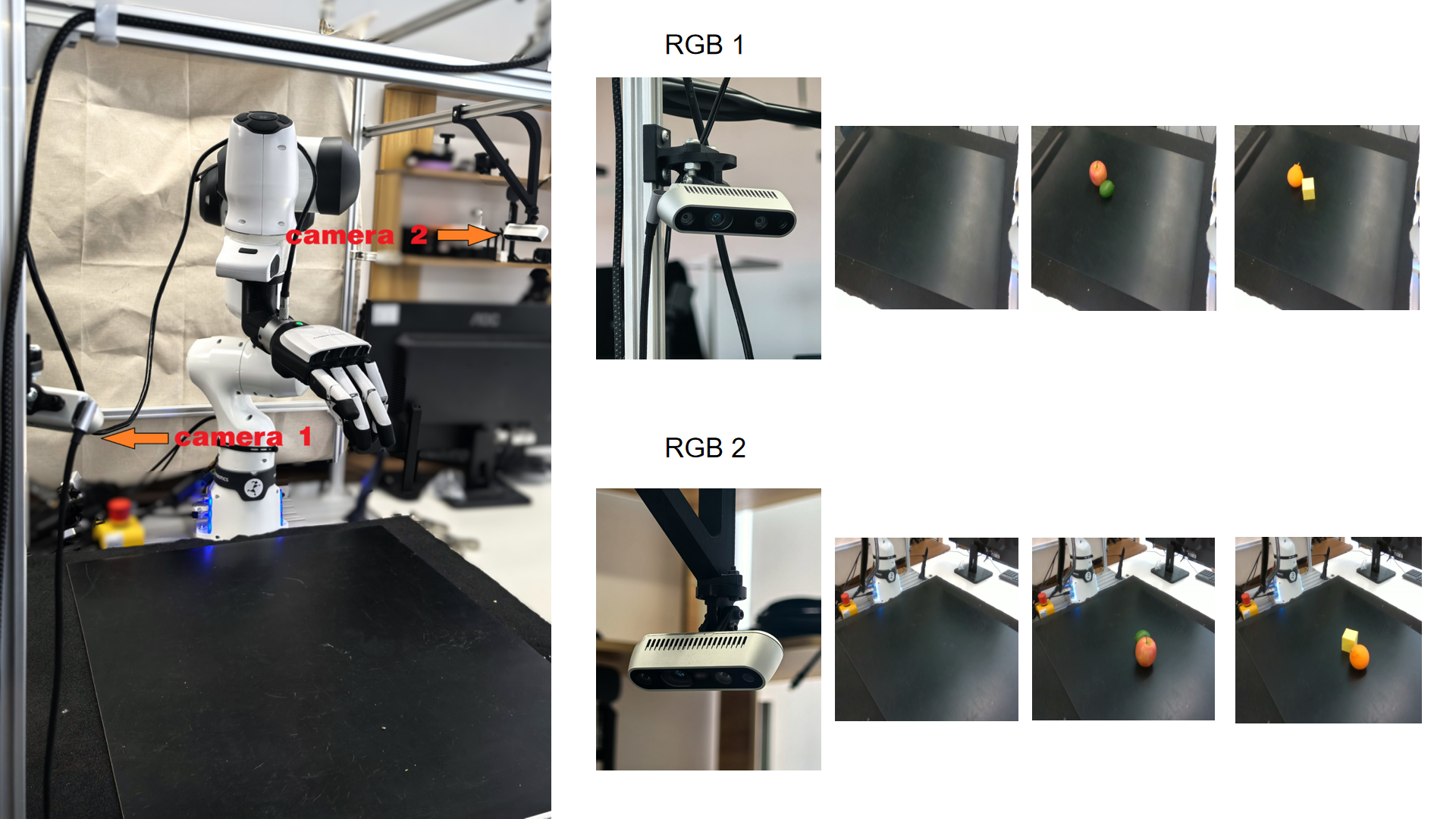}
    \caption{\textbf{Real-world hardware platform.} Left: Franka Research 3 robot arm equipped with an Inspire Hand and two fixed RGB cameras positioned at diagonal viewpoints around the workspace. Right: representative RGB observations used as policy inputs during deployment.
    }
    \label{fig:hardware_setup}
\end{figure}

We conduct real-world experiments using a 6-DoF Inspire Hand mounted on a 7-DoF Franka Research 3 robot arm. The world coordinate frame is defined with its origin at the robot base, and all object poses are expressed relative to this frame. Visual observations are captured by two fixed Intel RealSense D435i cameras positioned at diagonal viewpoints around the workspace, as shown in Figure~\ref{fig:hardware_setup}. Although the cameras provide RGB-D observations, only RGB images are used as input, and the depth information is discarded. Images from both cameras are resized to $256 \times 256$ pixels. This multi-view configuration mitigates self-occlusion during grasp execution and provides complementary views of the object interaction region. Policy inference is performed on a workstation equipped with an NVIDIA RTX 4090 GPU. The policy operates at approximately $3\mathrm{Hz}$. The control commands are executed directly on the physical system without tactile sensing, force--torque feedback, or auxiliary state-estimation modules. Thus, the real-world experiments rely exclusively on RGB observations and robot proprioception.

\subsection{Task Setup}
\begin{wrapfigure}{r}{0.42\linewidth}
\vspace{-0.6em}

\centering
\includegraphics[width=\linewidth]{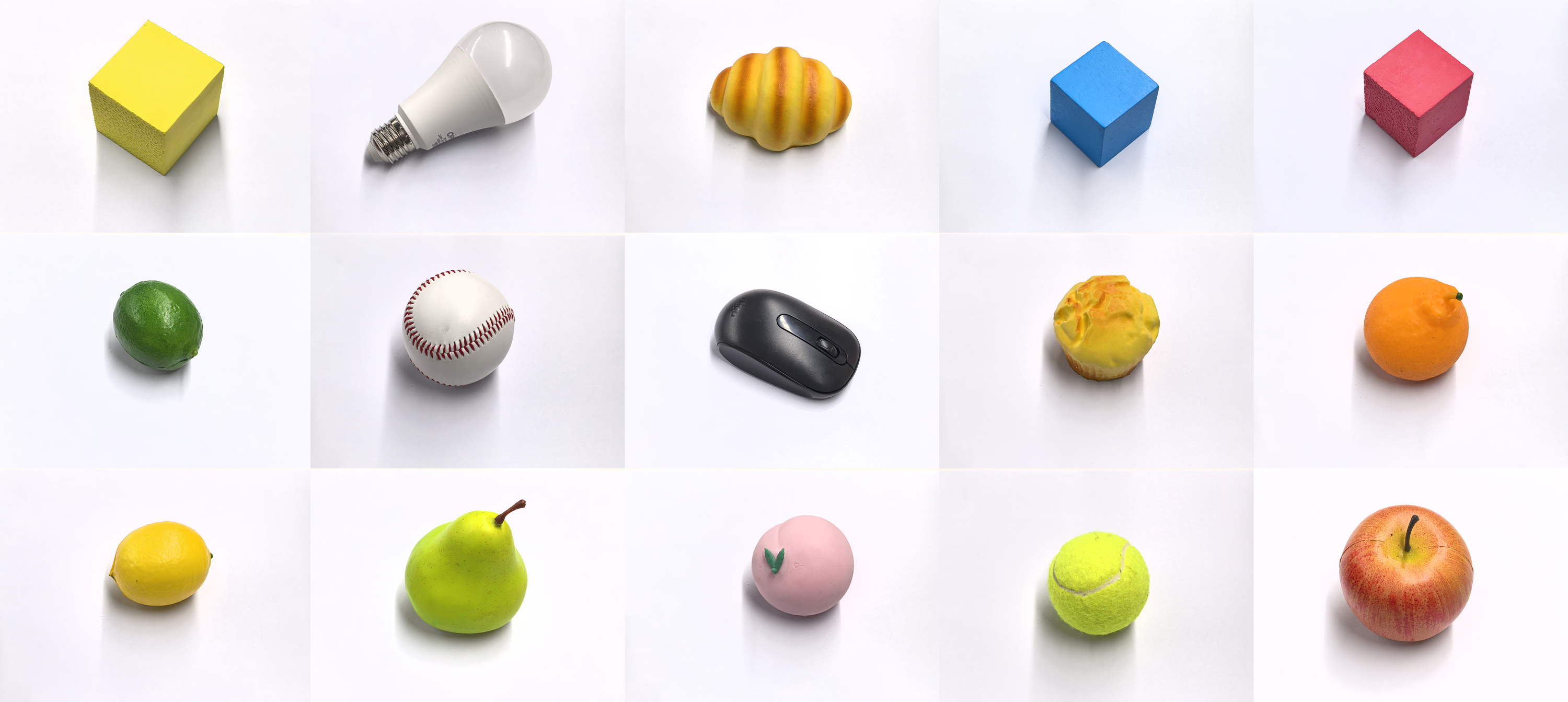}
\caption{
\textbf{Objects used in real-world evaluation.} Example objects that exhibit substantial diversity in geometry, size, texture, and material properties.}
\label{fig:sim2real_objects}
\vspace{-0.6em}
\end{wrapfigure}
Real-world evaluation is conducted on 20 everyday objects. All objects are strictly unseen during training and are selected to cover a diverse range of geometries, aspect ratios, surface materials, and visual appearances. Figure~\ref{fig:sim2real_objects} illustrates some objects used throughout the experiments. To construct the evaluation benchmark, we randomly sample 10 object-pair combinations. For each combination, 20 independent trials are performed, resulting in a total of 200 real-world evaluation episodes. A trial is considered successful only if both target objects are simultaneously lifted at least $10 \mathrm{cm}$ above their initial positions and remain stably grasped for at least 5 seconds. To evaluate robustness under real-world visual domain shifts, we additionally conduct qualitative experiments under six environmental configurations involving different background appearances and colored illumination conditions. These visual perturbations are intentionally introduced without any policy adaptation. While the quantitative success rate reported is measured under a standard evaluation setup, these additional experiments provide evidence of the policy's ability to maintain successful dual-object grasping behaviour under moderate visual variations.

\clearpage
\subsection{Detailed Sim-to-Real Results}

\subsubsection{Qualitative Results}
Figures~\ref{fig:sim2real_gallery} and~\ref{fig:sim2real_gallery_separated} present representative successful executions in proximal and separated layouts, respectively. The examples cover diverse object combinations with varying geometries, aspect ratios, and inter-object spatial arrangements. Across these cases, the policy demonstrates coordinated dual-object grasping and lifting without task-specific adaptation. These qualitative results provide further evidence that the grasping strategy, trained entirely in simulation, can transfer effectively to physical hardware despite discrepancies in visual appearance, object geometry, and contact dynamics.

\begin{figure}[H]
    \centering
    \includegraphics[width=\linewidth]{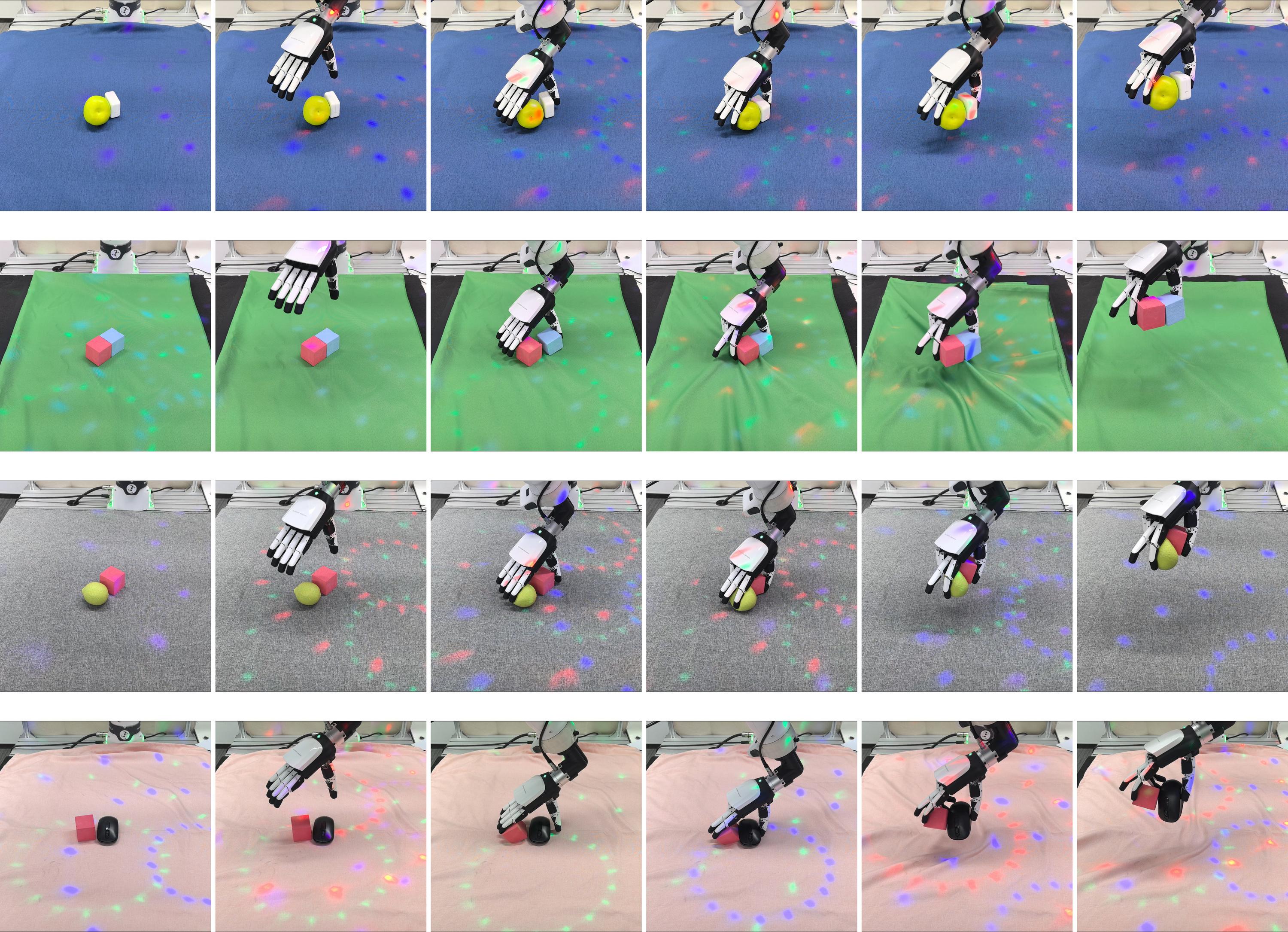}
    \setcounter{figure}{13}
    \caption{Real-world demonstrations in scenes with randomized backgrounds and lighting conditions.}
    \label{fig:sim2real_gallery_9527}
\end{figure}

To further assess robustness to real-world visual domain shifts, we deploy the policy under a range of visual conditions featuring different tablecloth textures, background colors, and lighting configurations. Representative execution trajectories are shown in Fig.~\ref{fig:sim2real_gallery_9527}. Across these appearance variations, the policy is able to perform successful dual-object grasping without additional adaptation or calibration. These qualitative results indicate that the visual representations learned during simulation training retain a degree of robustness to changes in environmental appearance, thereby supporting effective sim-to-real transfer.

\clearpage
\setcounter{figure}{11}
\begin{figure}[H]
    \centering
    \includegraphics[width=\linewidth,height=0.82\textheight]{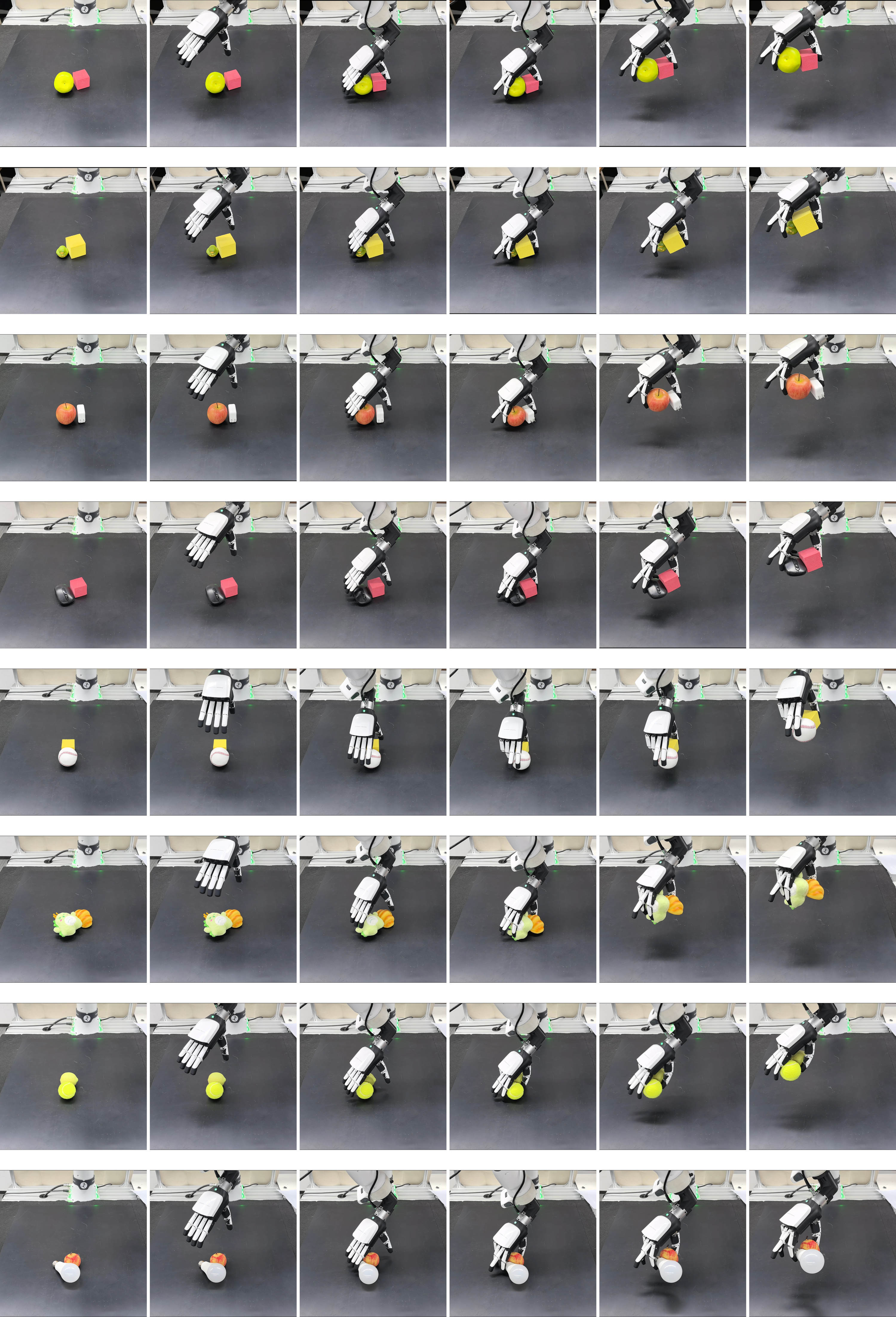}
    \caption{Real-world grasping trajectories in proximal layouts.}
    \label{fig:sim2real_gallery}
\end{figure}
\clearpage
\begin{figure}[H]
    \centering
    \includegraphics[width=\linewidth,height=0.82\textheight]{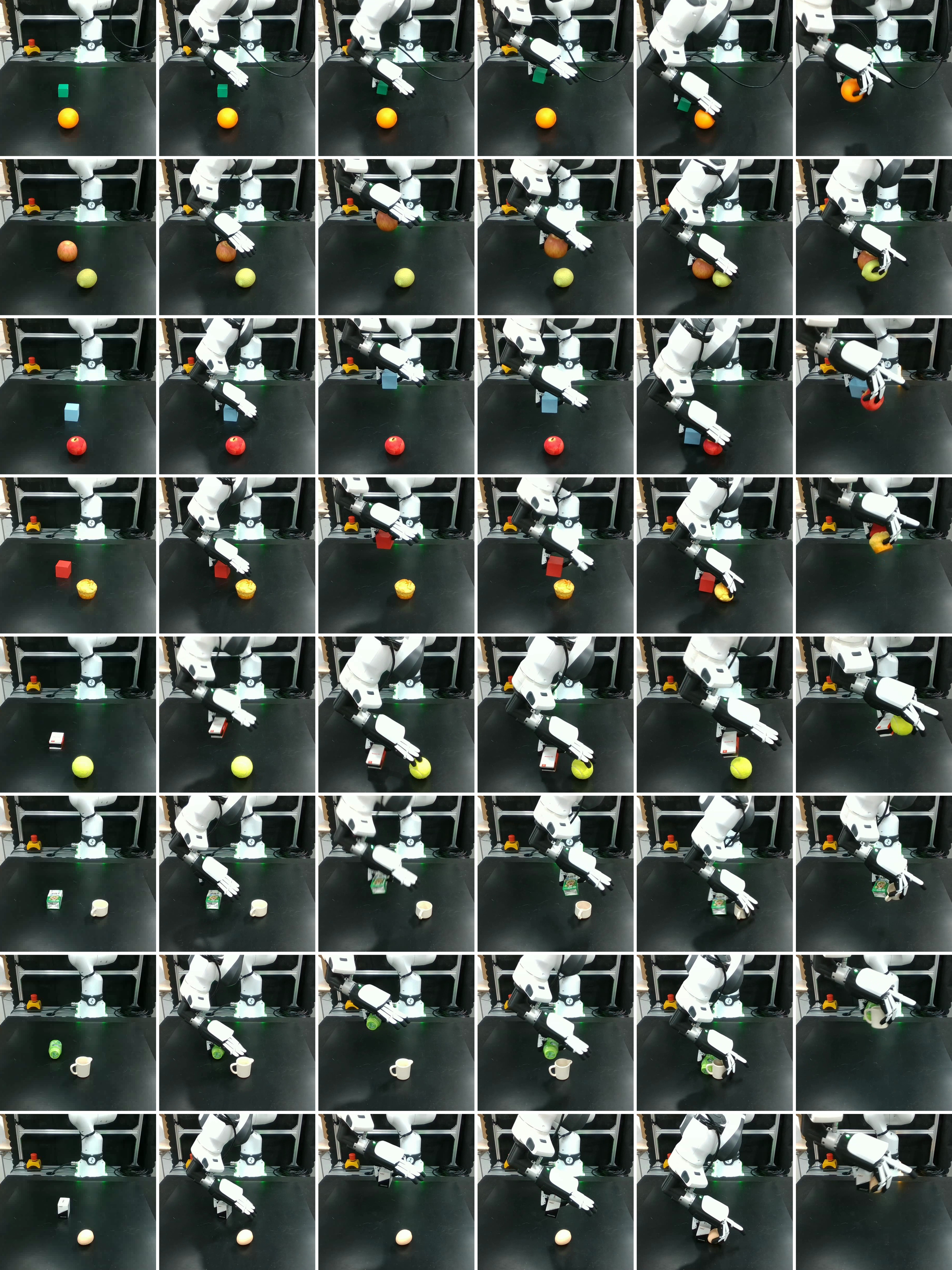}
    \caption{Real-world grasping trajectories in separated layouts.}
    \label{fig:sim2real_gallery_separated}
\end{figure}
\clearpage
\setcounter{figure}{14}
\subsubsection{Failure Case Analysis}

\begin{figure}[H]
    \centering
    \includegraphics[width=0.78\linewidth]{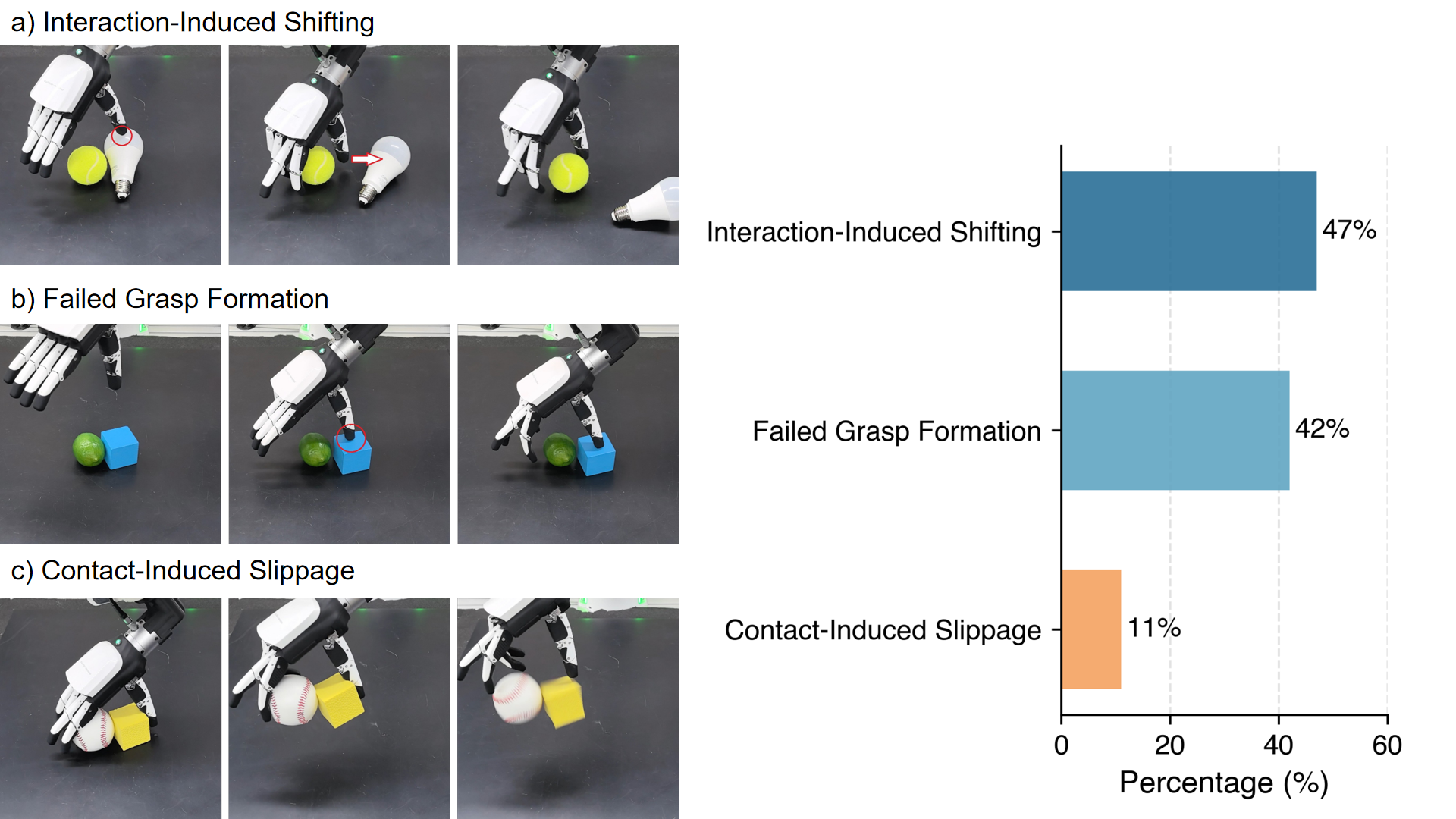}
    \caption{
    \textbf{Failure mode analysis in real-world deployment.}
    Left: representative examples of three dominant failure modes observed during real experiments. 
    (a) \textit{Interaction-Induced Shifting}: asymmetric contact causes one object to be displaced before a stable dual-object enclosure is formed.
    (b) \textit{Failed Grasp Formation}: the hand fails to simultaneously establish a stable enclosure around both target objects.
    (c) \textit{Contact-Induced Slippage}: one object slips from the grasp after successful lifting.
    Right: distribution of failure modes obtained from manual inspection of all failed real-world trials.
    }
    \label{fig:failed_cases}
\end{figure}

To better characterize the limitations of our vision-based policy, we manually analyze the failed real-world trials in proximal layouts and assign each trial to a failure category based on its primary cause. Figure~\ref{fig:failed_cases} shows representative examples of each category together with their overall distribution. Three dominant failure modes are identified: \textit{Interaction-Induced Shifting} (47\%), \textit{Failed Grasp Formation} (42\%), and \textit{Contact-Induced Slippage} (11\%).

Among the identified failure modes, \textit{Interaction-Induced Shifting} is the largest source of performance degradation. This failure typically occurs during finger closure, when perceptual uncertainty causes one object to be contacted slightly earlier than the other. The resulting asymmetric contact induces lateral object motion before a stable enclosure of both objects can be established. In many cases, one object is pushed out of the grasp region or becomes loosely wedged against the other, creating deviations that the open-loop policy cannot readily correct.

This phenomenon is further exacerbated by the temporal abstraction used by the VLA policy. From a single visual observation, the model predicts an action chunk comprising ten consecutive control commands, which are then executed open-loop for five steps before the next policy update. During this interval, small interaction-induced displacements may accumulate without corrective feedback. Moreover, severe occlusions caused by the hand frequently obscure the contact region, limiting the policy's ability to detect subtle relative motion between the objects.

\textit{Failed Grasp Formation} accounts for 42\% of the observed failures and refers to cases in which the hand fails to establish a stable enclosure of both objects before lifting. Typical examples include grasps that secure only one object and grasp configurations that are geometrically unsuitable for simultaneously enclosing both targets. These failures indicate that, although the policy can often generate feasible dual-object grasps, certain object arrangements remain challenging to be handled reliably. In contrast, \textit{Contact-Induced Slippage} accounts for only 11\% of the observed failures and occurs after both objects have been successfully lifted. In these cases, the hand initially establishes a stable enclosure of both objects, but one object gradually slips from the grasp as changes in load distribution reduce contact stability.

\end{document}